\PassOptionsToPackage{dvipsnames}{xcolor}
\PassOptionsToPackage{round,authoryear}{natbib}

\documentclass{article}
\usepackage[utf8]{inputenc}
\usepackage[T1]{fontenc}
\usepackage{main}
\usepackage{microtype}
\usepackage{graphicx}
\usepackage{times}
\usepackage{latexsym}
\usepackage{amsmath}
\usepackage{amssymb}
\usepackage{booktabs}
\usepackage{enumitem}
\usepackage{float}
\usepackage{xspace}
\usepackage{tabularx}
\usepackage{tcolorbox}
\usepackage{multirow}
\usepackage{adjustbox}
\usepackage{pgfplots}
\pgfplotsset{compat=1.16}
\usepackage{placeins}
\usepackage{alltt}
\usepackage{array}
\usepackage{makecell}
\definecolor{darkblue}{rgb}{0, 0, 0.5}
\definecolor{mydarkblue}{rgb}{0,0.08,0.45}
\usepackage[colorlinks=true,linkcolor=mydarkblue,citecolor=mydarkblue,filecolor=mydarkblue,urlcolor=mydarkblue]{hyperref}
\usepackage{fancyhdr}

\newcommand{\ours}{BackTrend\xspace}
\newcommand{\ntopic}{25\xspace}
\newcommand{\nweak}{66\xspace}
\newcommand{\nprob}{34\xspace}
\newcommand{\nsol}{32\xspace}

\definecolor{XiaoBlue}{rgb}{0.04,0.34,0.75}

\let\cite\citep

\definecolor{YaleBlue}{RGB}{0, 53, 107}
\definecolor{NYUpurple}{RGB}{87, 6, 140}
\definecolor{TCSblue}{RGB}{1, 126, 199}
\newcommand{\Yale}{\hspace{.1em}^{\textcolor{YaleBlue}{\boldsymbol{Y}}}}
\newcommand{\NYU}{\hspace{.1em}^{\textcolor{NYUpurple}{\boldsymbol{N}}}}
\newcommand{\TCS}{\hspace{.1em}^{\textcolor{TCSblue}{\boldsymbol{T}}}}
\newcommand{\huggingface}{\raisebox{-1.5pt}{\includegraphics[height=1.05em]{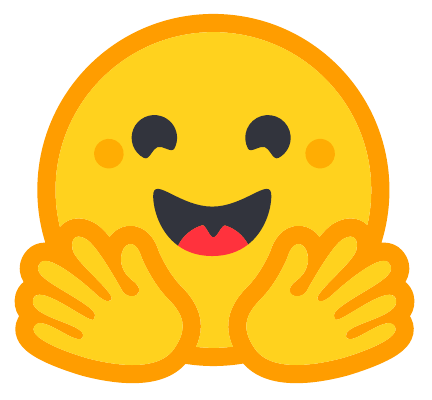}}\xspace}
\newcommand{\github}{\raisebox{-1.5pt}{\includegraphics[height=1.05em]{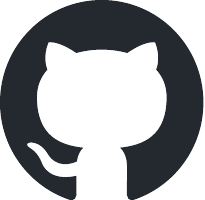}}\xspace}

\title{\fontsize{15}{17}\selectfont \ours: Evaluating Scientific Weak-Signal Prediction via \\ Backward Reconstruction}

\author{
\textbf{Xiao Zhou}$\Yale$\thanks{Equal contributions.} \quad
\textbf{Yilun Zhao}$\Yale$\footnotemark[1] \quad
\textbf{Owen Jiang}$\Yale$\footnotemark[1] \quad
\textbf{Tiansheng Hu}$\NYU$ \\ [5pt]
\textbf{Cai Xu}$\Yale$ \qquad
\textbf{Manasi Patwardhan}$\TCS$ \qquad
\textbf{Arman Cohan}$\Yale$ \\ [7pt]
$\Yale$Yale NLP Lab \qquad $\NYU$New York University \qquad $\TCS$TCS Research \\ [6pt]
\huggingface \href{https://huggingface.co/datasets/rebeccazzzz/BackTrend}{\ours Dataset}
\hspace{3em}
\github \href{https://github.com/rebeccaz4/BackTrend.git}{\ours Code}
}

\begin{document}

\fancyhf{}
\renewcommand{\headrulewidth}{0pt}
\renewcommand{\footrulewidth}{0pt}
\setlength{\headheight}{24pt}
\setlength{\headsep}{3pt}
\lhead{%
    \raisebox{0.00cm}{\includegraphics[height=0.52cm]{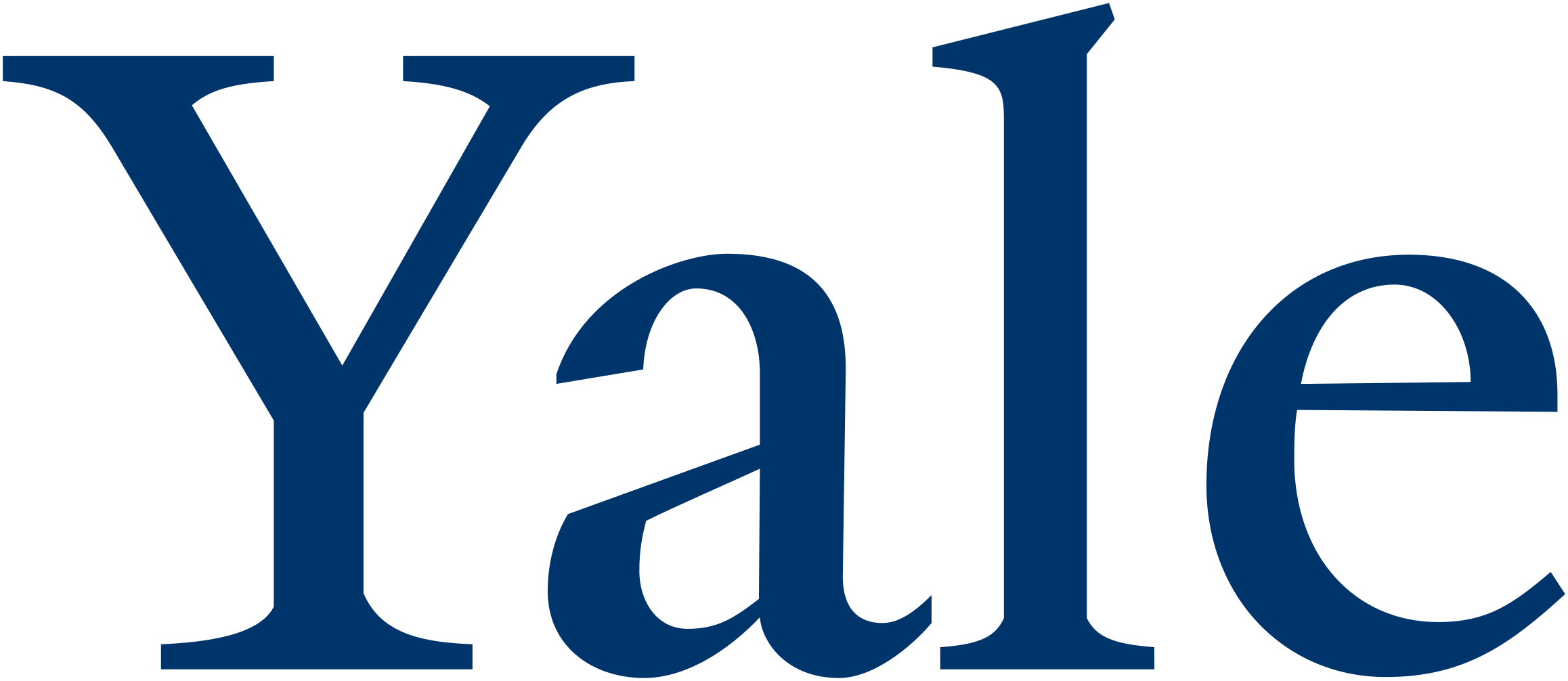}}\hspace{0.34cm}%
    \raisebox{-0.05cm}{\includegraphics[height=0.60cm]{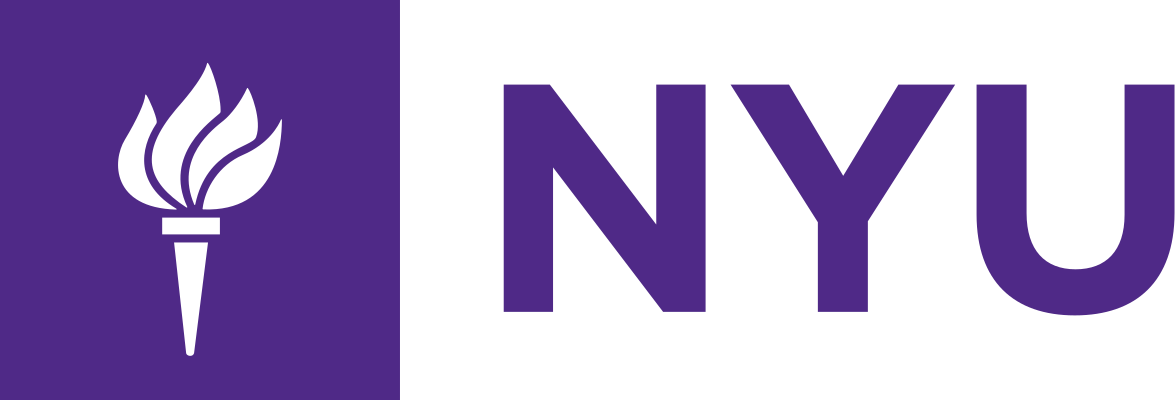}}%
}
\rhead{%
    \raisebox{0.00cm}{\includegraphics[height=0.50cm]{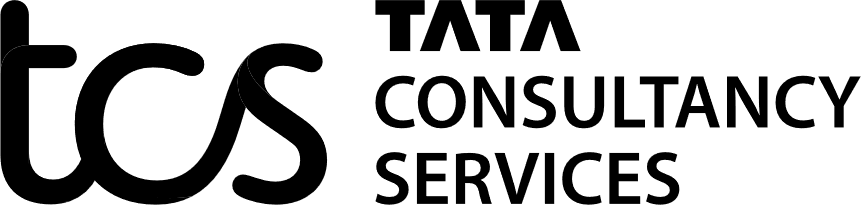}}%
}
\makeatletter
\renewcommand{\@toptitlebar}{%
  \vskip -0.35cm
  {\color{black}\hrule height 1\p@}
  \vskip 0.20in
  \vskip -\parskip%
}
\makeatother

\maketitle
\thispagestyle{fancy}
\pagestyle{plain}

\begin{abstract}
Scientific weak signals are early, low-visibility research directions that later become central to mature scientific topics, yet existing resources such as trend tracking, citation forecasting, and foresight reports rarely provide validated reference sets that link concrete early precursors to later paradigms. We introduce \ours, a retrospective benchmark in which, given a mature target topic and a temporal evidence constraint, systems must recover two types of precursors: \emph{problem-space signals}, underrecognized research problems, and \emph{solution-space signals}, emerging methods for known problems. \ours contains \ntopic mature target topics in artificial intelligence and machine learning and \nweak human-validated weak signals, reconstructed from large-scale literature by grounding each candidate in its 2019--2024 publication-frequency trajectory. We evaluate frontier LLMs, RAG systems, and agentic research systems using semantic matching and coverage-based metrics. Current systems often generate plausible but misaligned precursors, exhibiting topic drift, granularity mismatch, near-miss matching, and incomplete coverage; the strongest system achieves only 10.1\% F1, while Coverage\@$10$ reaches at most 18.5\% of the reference signals. Our budget analyses show that additional retrieval and web-search evidence can improve performance up to a moderate budget, but does not by itself close the substantial performance gap.

\end{abstract}

\section{Introduction}

\begin{figure}[tbp]
\centering
\includegraphics[width=\textwidth]{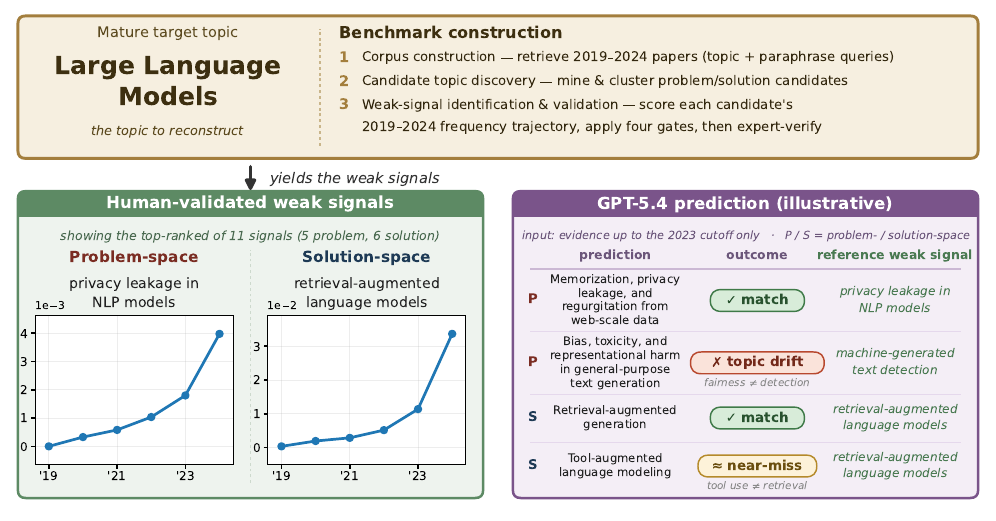}
\caption{\ours overview for the 2024 mature target topic \emph{large language models}. \textbf{Top:} the benchmark-construction pipeline (\S\ref{sec:direction_focused}), in which candidate \emph{problem-space} and \emph{solution-space} precursors are mined from the topic's 2019--2023 papers, scored by their 2019--2024 frequency trajectory, gated, and expert-validated. \textbf{Lower-left:} two of the topic's eleven human-validated weak signals (the top-ranked problem- and solution-space precursor), each with its 2019--2024 frequency trajectory, which shows the low-then-rising signature that defines a weak signal. \textbf{Lower-right:} four real GPT-5.4 predictions, verbatim from the prediction outputs and produced under a pre-2024 evidence cutoff, each listed with its outcome and with the reference weak signal it was scored against. The full error taxonomy, which also covers granularity mismatch and coverage failure, is defined in \S\ref{sec:error-analysis}.}
\label{fig:teaser}
\end{figure}

Identifying early indicators of transformative research, or weak signals, has long been a difficult goal across scientific disciplines \cite{ansoff1975weak, porter2004technologyfutures}. Detecting such signals is valuable for researchers, corporations, and policymakers because it informs long-term investment, research prioritization, and policy design \cite{kajikawaarticletrackingemerging, yoonarticledetectingsignals, leearticletheprioritization, OGAWA2015469}. As AI and machine learning accelerate the pace of scientific research, effective foresight depends on detecting low-visibility but potentially high-impact ideas within a large volume of background noise, and recent work continues to identify open issues in methods for detecting such emerging themes \cite{REN2021102196, LIU2023102872}. Across scientific domains, LLM-based research agents are increasingly deployed to search literature, synthesize emerging areas, and propose new methods \cite{zhao-etal-2026-rethinking, hu2026sage, yu2025alpharesearch}, and a growing body of benchmarks evaluates their scientific literature understanding and research capabilities \cite{zhao-etal-2025-abgen, xu-etal-2025-llms-identify, chen2026measuringgaphumanllm, zhao2026sciarena, wang-etal-2025-sciver, zhao-etal-2025-multimodal-foundation, chen-etal-2026-scimdr}, yet they are rarely tested on whether they can recover low-visibility precursors that later become central, leaving an open challenge for long-context reasoning and abstraction control.

We address this challenge with \textbf{\ours}, a retrospective benchmark for scientific weak-signal prediction (\autoref{fig:teaser}). Given a \emph{mature target topic} and a temporally restricted evidence window, systems must recover early \emph{problem-space} and \emph{solution-space} weak signals that later became important precursors of that topic. Because prospective evaluation would require waiting years for outcomes to mature, \ours uses backward reconstruction as a controlled proxy for scientific foresight. We construct the reference set by reconstructing candidate precursors directly from large-scale Semantic Scholar literature: for each mature target topic, we mine candidate research directions from its 2019--2023 papers and score them by their 2019--2024 publication frequency, so that a candidate is retained only when a low-visibility early emergence is followed by a clear rise in its 2024 frequency. AI/ML researchers among the authors then validate every retained candidate for topical relevance, temporal consistency, and supporting evidence; at evaluation time, systems must recover the held-out signals using only evidence from publications available before the end-of-2023 prediction cutoff.

Our results show that strong systems can produce plausible precursors but routinely miss the benchmark's intended structure. Across all seven evaluated systems, semantic-judge F1 and Coverage\@$K$ remain low: F1 reaches at most 10.1\% Coverage\@$K$ reaches at most 18.5\%. The dominant errors are topic drift, granularity mismatch, lexical near-miss, and coverage failure. Additional evidence can improve performance, but not uniformly: deeper retrieval generally benefits both RAG models, with a stronger effect on Qwen3-8B, while DeepResearch performs best with a moderate search budget rather than unlimited search. Weak-signal prediction therefore tests semantic alignment, abstraction-level control, and coverage over concrete research precursors rather than fluency in generating scientific phrases.

Our main contributions are summarized below:
\begin{itemize}[leftmargin=*]
\itemsep0em
\item We formulate \emph{retrospective scientific weak-signal prediction}: recovering early problem-space and solution-space precursors of mature research topics under a temporal evidence constraint that simulates research foresight.
\item We construct \ours, a benchmark of \ntopic mature target topics in artificial intelligence and machine learning and \nweak human-validated weak signals, reconstructed from large-scale literature by scoring each candidate's 2019--2024 publication-frequency trajectory and verified by AI/ML researchers among the authors. 
\item We benchmark frontier LLMs, RAG systems, and DeepResearch agents on \ours, finding that current systems produce plausible but misaligned signals and that semantic-judge F1 reaches at most 10.1\% for every system and Coverage@$10$ at most 18.5\%, with errors dominated by topic drift, granularity mismatch, lexical near-miss, and coverage failure.
\end{itemize}

\section{Related Work}
\paragraph{Benchmark Construction for Emerging Topics and Weak Signals.}
Existing benchmarks for emerging-topic analysis mainly evaluate whether systems can detect and track salient events or trends over time, rather than identify early scientific ideas that later develop into dominant paradigms \cite{bookjamesallantopicdetection, petrovic-etal-2010-streaming, deng2022title2eventbenchmarkingopenevent}. In scientific domains, benchmark construction has also focused on trajectory labeling or future impact prediction, such as classifying topics as rising or declining or forecasting citations \cite{prabhakaran-etal-2016-predicting, Moiseeva_Schütze_2020, ofer2023forecasting, gu2024impactcast, ajith2026presciencebenchmarkforecastingscientific}. Foresight and horizon-scanning reports offer another related resource, but the signals they curate are usually broad thematic areas rather than concrete technical precursors \cite{day2007weak, Saritas2013, 10.3152/030234210X484801}. \ours addresses these limitations by constructing benchmark instances retrospectively from mature research topics, defining weak signals as empirically grounded early precursors rather than transient events, broad themes, or short-term impact patterns. Each instance decomposes into problem-space and solution-space precursors and is evaluated under a prediction cutoff that withholds post-cutoff information from the system.

\paragraph{Weak Signal Detection and Emerging Trend Analysis.}
Weak-signal detection has long been studied in foresight, bibliometrics, and technology intelligence as the task of identifying faint early indicators of future change \cite{ansoff1975weak, hiltunen2008, holopainen2012}. Early computational approaches relied on keyword frequencies, citation networks, and co-occurrence statistics, while later work adopted topic models and contextual embedding methods to better capture latent semantic structure and thematic evolution over time \cite{small1973, yoon2012, song2017, blei2003lda, rudolph2018, yao2018, grootendorst2022bertopic, boutaleb2024bertrendneuraltopicmodeling, Ebadi_2026}. However, these methods are still typically evaluated through case studies or proxy trend metrics, because existing resources rarely provide verified examples of concrete early precursors to mature scientific themes. \ours addresses this gap with a precursor-recovery task featuring validated targets and a unified protocol. We benchmark LLM-based research agents and omit traditional bibliometric or topic-modeling systems because their outputs are not directly aligned with \ours’s output format: given a mature topic, \ours requires an explicit, topic-conditioned set of precursor hypotheses, whereas these methods typically produce corpus-level trends, clusters, or topic trajectories. Adapting them would require an additional, non-standardized mapping from such outputs to target-specific precursor labels, potentially confounding direct comparison. We leave such adaptations to future work.  %

\begin{figure}[tbp]
\centering
\includegraphics[width=\textwidth]{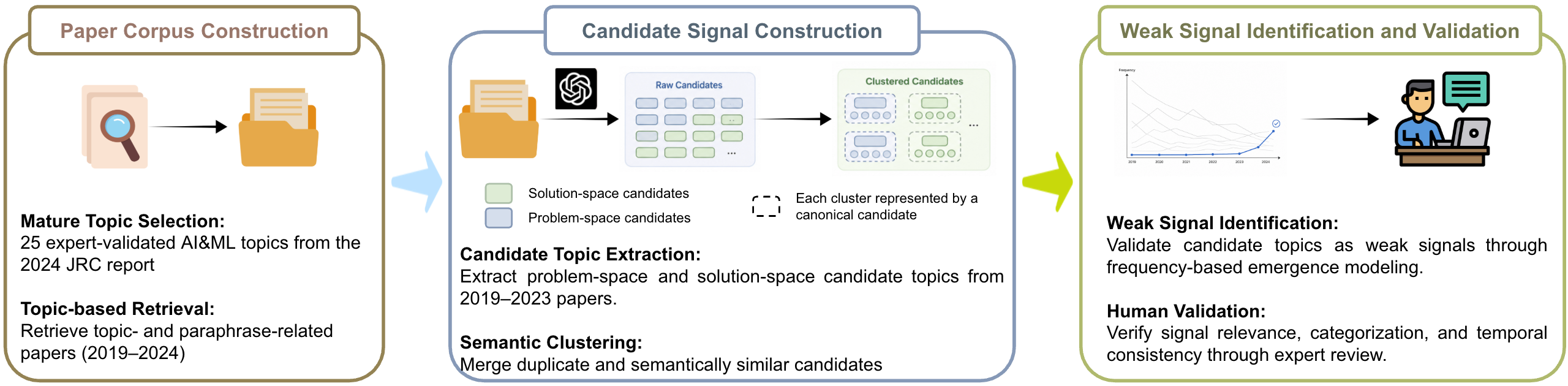}
\caption{Overview of the \ours construction pipeline. We first compile \ntopic{} mature target topics from the Artificial Intelligence and Machine Learning domain of a 2024 JRC report~\cite{jrc2024weaksignals} and retrieve their 2019--2024 papers from Semantic Scholar. We then mine problem-space and solution-space candidate topics from the 2019--2023 abstracts, consolidate them by embedding clustering, and score each candidate by its 2019--2024 publication frequency, keeping only candidates that pass all four frequency gates. Finally, two AI/ML researchers among the authors validate the surviving candidates for topical relevance, correct problem/solution categorization, temporal consistency, and evidence faithfulness to produce the final \ours benchmark.}
\label{fig:pipeline}
\end{figure}

\section{\ours Benchmark}
\label{headings}
\ours is a retrospective weak-signal prediction benchmark in which systems recover early problem-space and solution-space precursors of mature research topics. Below we define the task and the two weak-signal categories (\S\ref{sec:backtrend-task}), detail the construction pipeline with expert validation (\S\ref{sec:direction_focused}), and report per-topic statistics (\S\ref{sec:data-stats}).

\subsection{\ours Task}
\label{sec:backtrend-task}
The \ours{} task frames scientific foresight as recovering a mature topic’s early, low-visibility precursors from evidence available before it matured, using backward reconstruction as a controlled proxy for prediction. We define a precursor as an early research direction that later becomes associated with or central to the mature topic, without implying a causal relationship.

\paragraph{Mature Target Topics.}
Given a scientific domain, mature target topics serve as anchor concepts for weak-signal construction. Each mature target topic represents a research direction that has already achieved substantial visibility within its domain through sustained publication activity. For example, \emph{large language models} constitute a mature target topic in 2024.
In \ours, these topics are instantiated from externally validated AI and machine-learning topics in the 2024 JRC weak-signal report \cite{jrc2024weaksignals}; we describe the selection process in \S\ref{sec:direction_focused}.

\paragraph{Problem-space and Solution-space Weak Signals.}
We define a weak signal as a concrete research direction that initially received limited attention but later grew exponentially in prominence. We distinguish two types: (1) \emph{problem-space weak signals} are underrecognized research problems or problem formulations that later become central to the mature target topic in a given field; (2) \emph{solution-space weak signals} are early or niche methods, techniques, or design principles that were not yet widely adopted but later became important solutions to already recognized problems. In short, problem-space signals surface new questions, while solution-space signals offer emerging answers to existing ones.

\paragraph{Task Formulation.}
We formulate weak-signal discovery as the task of identifying historically grounded early research directions that later contribute to the emergence of a given mature target topic. Formally, given a domain-specific mature target topic $M$, a maturity year $y$, an evidence window $t=[y-k,y-1]$, and a designated signal space (problem or solution), the model must retrieve or infer a set of candidate weak signals $s=\{s_1,\dots,s_n\}$. A valid weak signal $s_i \in s$ must (i) exhibit low visibility during its early stage, (ii) demonstrate growth in prominence within the evidence window, and (iii) have this rise confirmed in the maturity-year corpus of $M$, with its frequency in year $y$ exceeding its highest frequency within the evidence window by a factor of at least $\lambda=1.2$ (\S\ref{sec:direction_focused}), where $\lambda$ denotes the maturity-year frequency lift threshold. The year $y-1$ is referred to as the \emph{prediction cutoff year}: at evaluation time, systems are given the mature target topic $M$ but may only use evidence available up to the end of this year, simulating a research-foresight setting in which evidence from the maturity year about how $M$ emerged is unavailable to the model. In \ours, mature topics are drawn from the 2024 JRC report \cite{jrc2024weaksignals}, where $y=2024$ and $k=5$, resulting in an evidence window of 2019--2023 and a prediction cutoff year of 2023. 
\subsection{Benchmark Construction}
\label{sec:direction_focused}

We next detail the \ours construction process, with an overview shown in \autoref{fig:pipeline}. The pipeline is grounded in observed publication frequency: a candidate is retained only when a low-visibility, exponentially growing early emergence is followed by a rise in its frequency in the 2024 mature-topic corpus.

\paragraph{Corpus Construction.}
To identify weak signals retrospectively, we begin with a set of research topics that have already reached maturity. We obtain these mature topics from the 2024 JRC weak-signal report~\cite{jrc2024weaksignals}, which identifies science and technology topics through large-scale data analysis and expert validation. Specifically, we focus on topics within the Artificial Intelligence and Machine Learning domain and backtrack their historical development to uncover the weak signals that preceded them. This results in a benchmark comprising 25 mature topics spanning a diverse range of AI and machine learning research directions. For each mature target topic $M$, we retrieve papers through the Semantic Scholar API using keyword queries derived from the topic name and its manually curated paraphrases, with one query per year from 2019 to 2024, to capture alternative terminology and improve retrieval coverage. These paraphrase queries expand corpus coverage and reduce the risk of missing historically relevant papers. The retrieved papers form the topic/paraphrase paper corpus, whose members we refer to as $M$-papers. Papers from 2019--2023 serve as the historical corpus for candidate-topic discovery, while 2024 papers provide the later-year frequencies used to test whether an early candidate's frequency subsequently rose. 

\paragraph{Candidate Topic Discovery.}
We next extract candidate topics from the 2019--2023 historical corpus. For each paper abstract retrieved for a mature target topic, we extract up to two reusable literature-level candidate topics that are explicitly grounded in the abstract and conceptually related to the target topic; the extraction prompts are shown in \autoref{fig:construct-prompt-a} and \autoref{fig:construct-prompt-b}. Each candidate is assigned to one of two spaces: \emph{problem-space} candidates describe research problems, limitations, risks, gaps, evaluation failures, or scientific questions, whereas \emph{solution-space} candidates describe reusable methods, system directions, defenses, benchmarks, datasets, or evaluation protocols. We require candidate labels to be neither overly broad field names nor paper-specific implementation details, and we avoid problem-solution phrases that conflate a method with the problem it addresses.

The extracted candidates are then consolidated within each mature target topic and candidate type. We first exact-deduplicate normalized candidate strings while retaining their supporting source-paper IDs, years, evidence snippets, and mention counts. We then encode candidate-topic strings with text-embedding-3-large\footnote{https://platform.openai.com/docs/guides/embeddings} and cluster semantically similar candidates using a cosine-similarity threshold of 0.85, chosen to balance semantic consolidation of related candidate topics with separation of distinct research directions. Clustering is performed separately for problem-space and solution-space candidates to avoid merging research problems with solution methods. For each cluster, we select a canonical candidate-topic label based on supporting evidence, prioritizing candidates with more source papers and mentions while favoring compact labels when support is comparable. The resulting clustered candidate topics are the units considered in the final weak-signal identification step.

\paragraph{Weak Signal Identification and Validation.}
For each clustered candidate topic $c$ of mature target topic $M$, we compute its yearly frequency over 2019--2024. Let $n_y(c)$ be the number of non-survey source papers supporting $c$ in year $y$, and $N_y(M)$ the number of non-survey papers retrieved for $M$ in year $y$; the frequency is
\[
f_y(c;M)=\frac{n_y(c)}{N_y(M)}.
\]
We exclude survey papers from both counts: a single survey touches many topics at once and would spuriously inflate a candidate's frequency, so excluding surveys reduces false-positive matches. However, we retain them during candidate discovery and clustering because surveys provide broad coverage of research directions. We further validate this design choice in Appendix~\ref{app:surveys}. 

For 2019--2023, a paper supports $c$ when its abstract matches $c$'s cluster. For 2024, we instead count a paper as supporting $c$ when it \emph{cites} at least one of $c$'s early 2019--2023 source papers:
\[
f_{2024}(c;M)=\frac{r_{2024}(c)}{N_{2024}(M)},
\]
where $r_{2024}(c)$ is the number of such non-survey 2024 $M$-papers. We measure the 2024 frequency through citations rather than by re-matching $c$'s wording because a research direction is often renamed or rephrased as it matures, and a direct text or embedding match in 2024 would miss these drifted mentions; a citation to the candidate's own early papers tracks the same line of work however it is now phrased. 

A candidate is selected only if it passes four gates $g_1$--$g_4$. Let $\tau_c\in\{2019,\dots,2022\}$ be its onset year, selected as the maximizer of the growth score below, and $\epsilon$ a small constant. The gates use four hyperparameters: a 2024-frequency lift $\lambda=1.2$, a year-to-year retention $\rho=0.8$, a pre-onset tolerance $\delta=0.6$, and a sparse-year skip $\kappa=1$. We first define the auxiliary quantities

{\small
\begin{align*}
F_{\max}  &= \textstyle\max_{2019\le y\le 2023}\, f_y(c;M), \\
F_{\tau}  &= \textstyle\max_{\tau_c\le y\le 2023}\, f_y(c;M), \\
F_{<\tau} &= \textstyle\max_{y<\tau_c}\, f_y(c;M), \\
A_y(c)    &= \mathbf{1}\big[\,f_{y+1}(c;M)+\epsilon\ge\rho\, f_y(c;M) \\
          &\qquad\ \ \vee\ \ n_y(c)\le\kappa\,\big],
\end{align*}
}
The growth score in $g_1$ fits an exponential trend to the frequencies over $[\tau_c,2023]$ and rewards a well-fitted, sustained rise from a low base, with $\tau_c$ the onset year that maximizes it; Appendix~\ref{app:score} gives the full definition. We then require all four gates to hold:
\[
\small
\begin{aligned}
g_1(c)&=\mathbf{1}[\mathrm{score}(c)>0],\\
g_2(c)&=\mathbf{1}\!\left[
f_{2024}(c;M)+\epsilon\ge \lambda F_{\max}
\right],\\
g_3^d(c)&=
\begin{cases}
\mathbf{1}[f_{2023}(c;M)+\epsilon\ge \rho F_{\tau}],
& d=\mathrm{prob},\\
\mathbf{1}\!\left[\prod_{y=\tau_c}^{2022} A_y(c)=1\right],
& d=\mathrm{sol},
\end{cases}\\
g_4(c)&=\mathbf{1}\!\left[
F_{<\tau}\le
\max\!\left(f_{\tau_c}(c;M),\delta f_{2023}(c;M)\right)
\right].
\end{aligned}
\]
The four gates operationalize the three clauses of the definition in \S\ref{sec:backtrend-task}, each ruling out one failure mode. $g_1$ requires the onset-window trajectory to fit a rising exponential, ruling out flat, declining, and single-point traces. $g_2$ requires the rise to be confirmed in the maturity year, ruling out candidates whose 2024 frequency never clears their own early peak. $g_3$ requires the growth to persist to the cutoff rather than collapse along the way, ruling out candidates that spike early and then fade. $g_4$ requires the candidate to have been faint before its onset year, ruling out directions that were already established when they began to grow. Together, $g_1$, $g_3$, and $g_4$ characterize the observable early-stage properties of a weak signal, while $g_2$ validates whether the signal eventually contributes to the mature topic.

A candidate is retained when $g_1(c)g_2(c)g^d_3(c)g_4(c)=1$, with $d=\mathrm{prob}$ for problem-space candidates and $d=\mathrm{sol}$ for solution-space candidates. After this automatic filtering step, two AI/ML researchers among the authors conduct human validation: they inspect the candidate label, source papers, its 2024 citation evidence, and frequency trajectory, then verify topical relevance, correct problem/solution categorization, temporal consistency, and evidence faithfulness. Both experts independently label all 125 gate-passing candidates as valid weak signals or not, with an inter-annotator agreement of Cohen's $\kappa=0.75$. Candidates that fail validation are revised or removed through consensus adjudication, yielding the final \nweak{} signals.

Appendix~\ref{app:running-example} traces a single candidate, \emph{retrieval-augmented language models} under the mature target topic \emph{large language models}, through every stage of the pipeline as a running example.

\subsection{Data Statistics}
\label{sec:data-stats}

\ours covers \ntopic mature target topics and \nweak human-validated weak signals (\nprob problem-space and \nsol solution-space), all within the Artificial Intelligence and Machine Learning domain of the 2024 JRC report. Per-topic statistics are reported in \autoref{tab:domain_summary}: the \nweak signals are distributed over 18 of the \ntopic topics, while the remaining seven yield no candidate that survives both the four gates of \S\ref{sec:direction_focused} and expert validation; \S\ref{app:absent} traces each empty set to an identifiable property of the topic's literature. At evaluation time, systems are asked to recover the weak signals associated with each mature target topic under an end-of-2023 temporal search cut-off, using only historically available information, simulating prospective foresight.

  \begin{table}[htbp]
  \centering
  \footnotesize
  \setlength{\tabcolsep}{5pt}
  \renewcommand{\arraystretch}{1.05}
  \resizebox{\linewidth}{!}{%
  \begin{tabular}{@{}lrrr@{\hspace{2.2em}}lrrr@{}}
  \toprule
  \textbf{Mature Target Topic} & \textbf{\#\,Papers} & \textbf{\#\,P-WS} & \textbf{\#\,S-WS} & \textbf{Mature Target Topic} & \textbf{\#\,Papers} & \textbf{\#\,P-WS} & \textbf{\#\,S-WS} \\
  \midrule
  Artificial Intelligence of Things      &  25.8K &  1 &  0 & Machine Unlearning                     &   1.9K &  2 &  2 \\
  Asynchronous Federated Learning        &   3.0K &  2 &  1 & Masked Face Recognition                &   1.9K &  1 &  1 \\
  Attention Mechanisms in CNN            &  29.5K &  3 &  5 & Masked Language Model                  &   5.4K &  0 &  3 \\
  Decentralized Federated Learning       &   4.3K &  1 &  1 & Multimodal AI                          &   8.6K &  0 &  0 \\
  Epistemic AI                           &   1.2K &  0 &  0 & Multimodal Hate Speech                 &   0.1K &  0 &  0 \\
  Evolutionary Neural Arch.\ Search      &   1.1K &  1 &  0 & Privacy-Preserving Machine Learning    &  11.3K &  1 &  0 \\
  Explainable AI                         &  18.8K &  2 &  1 & Scientific Machine Learning            &  35.4K &  0 &  1 \\
  Federated Deep Learning                &   5.5K &  0 &  0 & Self-Supervised CNN                    &   2.6K &  0 &  0 \\
  Federated Machine Learning             &   6.8K &  4 &  6 & Tiny Machine Learning                  &   3.6K &  1 &  0 \\
  Federated Reinforcement Learning       &   1.6K &  0 &  0 & Trustworthy AI                         &  29.1K &  4 &  2 \\
  Human--AI Interface                    &  22.4K &  1 &  1 & Trustworthy Machine Learning           &  26.6K &  0 &  0 \\
  Human-Centric AI                       &   7.2K &  1 &  1 & Vertical Federated Learning            &   1.0K &  4 &  1 \\
  Large Language Models                  &  22.8K &  5 &  6 &                                        &        &    &    \\
  \midrule
  \multicolumn{5}{@{}l}{\textbf{Total} \emph{(grand total across all \ntopic{} topics)}} & \textbf{277.7K} & \textbf{34} & \textbf{32} \\
  \bottomrule
  \end{tabular}%
  }
    \caption{Data statistics of \ours{}. All \ntopic{} mature target topics
  fall in the Artificial Intelligence and Machine Learning domain.
  \textbf{\#\,Papers}: unique non-empty-abstract papers retrieved for the topic and its
  paraphrases over 2019--2024 (thousands; entries are rounded to 0.1K and need not sum
  exactly to the total).
  \textbf{\#\,P-WS} / \textbf{\#\,S-WS}: number of human-validated problem- /
  solution-space weak signals. Seven topics yield no validated signal: five produce
  no gate-passing candidate, and for two, expert validation removes every
  gate-passing survivor (\S\ref{app:absent}). }
  \label{tab:domain_summary}
  \end{table}

\section{Experiment Setup}
We next discuss the evaluated systems and our automated and human evaluation protocols.
\subsection{Evaluated Systems}
\label{sec:evaluated-systems}
We evaluate three categories of weak-signal prediction systems. All systems operate under the end-of-2023 prediction cutoff defined in \S\ref{sec:backtrend-task}, enforced through prompt instructions for parametric systems and through hard filtering of retrieval results for systems that can search external corpora at inference time.

\paragraph{Frontier LLMs.}
We include GPT-5.4, Qwen3.5-397B-A17B (Qwen3.5-397B)~\cite{qwen3.5}, and DeepSeek-R1-0528~\cite{deepseekai2025deepseekr1incentivizingreasoningcapability}. Temporal consistency is enforced via explicit prompt instructions specifying the emergence time of weak signals.

\paragraph{Retrieval-Augmented LLMs (RAG).}
We evaluate Qwen3-8B and Qwen3-30B \cite{qwen3technicalreport} with retrieval augmentation, which are equipped with retrieval mechanisms to better ground predictions in evidence from historical scientific literature. For each mature target topic, we retrieve a set of candidate papers from Semantic Scholar using the topic as a query. The retrieved papers are encoded using the bge-base-en-v1.5 embedding model \cite{bge_embedding} and ranked by cosine similarity. The top 50 most relevant papers are then selected and provided as external context to support weak-signal prediction. To ensure temporal consistency, retrieved papers are filtered on the client side to include only those published on or before the end of 2023. We deliberately keep this retrieval stage simple; stronger instruction-following or reasoning-intensive retrievers and rerankers \cite{song-etal-2025-ifir, song-etal-2025-limrank, zhao-etal-2026-rethinking}, as well as alternative ways of constructing retrieval queries when the mature target topic is unavailable, could be explored in future work. 

\paragraph{DeepResearch Systems.}
We evaluate two agentic LLM systems, DR-Tulu-8B~\cite{shao2025dr} and Tongyi-DeepResearch-30B-A3B (Tongyi-DR-30B-A3B)~\cite{tongyidr}, which integrate multi-step reasoning with tool-augmented academic retrieval and evidence collection. Tongyi-DR-30B-A3B follows a ReAct-style framework, where the model iteratively performs reasoning and tool calls to query Semantic Scholar and refine its predictions. In contrast, DR-Tulu adopts a workflow-based agentic search pipeline that orchestrates model inference and retrieval through a modular multi-service architecture. In both systems, given a mature target topic, the model retrieves relevant papers and generates weak signals grounded in the collected academic evidence. To ensure temporal consistency, retrieval is restricted to papers published on or before the end of 2023, preventing information leakage from future publications during the multi-step search process.

\begin{table}[t]
      \centering
      \small
      \setlength{\tabcolsep}{4.5pt}
      \renewcommand{\arraystretch}{1.0}
      \begin{tabular}{l ccc ccc ccc c}
      \toprule
      \multirow{2}{*}{\textbf{Model}} & \multicolumn{3}{c}{\textbf{Set LLM}} & \multicolumn{3}{c}{\textbf{Signal LLM}} & \multicolumn{3}{c}{\textbf{Cov@10}} & \multirow{2}{*}{\textbf{Human}} \\
      \cmidrule(lr){2-4}\cmidrule(lr){5-7}\cmidrule(lr){8-10}
      & \textbf{Problem} & \textbf{Solution} & \textbf{All} & \textbf{Problem} & \textbf{Solution} & \textbf{All} & \textbf{Problem} & \textbf{Solution} & \textbf{All} & \\
      \midrule
      \multicolumn{11}{@{}l}{\emph{Frontier LLMs}} \\
      GPT-5.4            & 7.4 & 3.6 & 5.7 & 7.0 & 1.6 & 4.5 & 14.3 & 14.5 & 14.4 & 6.0 \\
      Qwen3.5-397B       & 6.5 & 4.0 & 5.3 & 4.7 & 1.1 & 3.0 & 7.5 & 9.5 & 8.4 & 5.3 \\
      DeepSeek-R1-0528   & \textbf{14.1} & \textbf{5.4} & \textbf{10.1} & \textbf{9.8} & \textbf{4.1} & \textbf{7.1} & \textbf{23.2} & 13.1 & \textbf{18.5} & \textbf{9.0} \\
      \midrule
      \multicolumn{11}{@{}l}{\emph{Retrieval-augmented LLMs}} \\
      Qwen3-30B (RAG)    & 9.3 & 3.5 & 6.6 & 6.7 & 3.1 & 5.0 & 10.1 & 9.5 & 9.8 & 8.0 \\
      Qwen3-8B (RAG)     & 6.5 & 2.1 & 4.4 & 5.6 & 2.1 & 4.0 & 13.3 & 11.9 & 12.7 & 5.0 \\
      \midrule
      \multicolumn{11}{@{}l}{\emph{DeepResearch agents}} \\
      Tongyi-DR-30B-A3B  & 6.7 & 4.3 & 5.6 & 5.2 & 3.2 & 4.3 & 14.1 & 6.2 & 10.4 & 6.0 \\
      DR-Tulu-8B         & 9.0 & 2.5 & 6.0 & 4.0 & 2.2 & 3.2 & 13.6 & \textbf{15.2} & 14.4 & 6.9 \\
      \bottomrule
      \end{tabular}
      \caption{Main results over the 30 topic-direction pairs with validated weak signals,
  contributed by 18 of the \ntopic{} AI/ML target topics. We report set-level, signal-level LLM-judge F1 (Claude-Opus-4.8, 3 runs) and Coverage@10 (Cov@10), broken down by signal direction (Problem / Solution / All), together with human-judged F1 (Human) from our annotation interface. The highest score in each column is in bold.
      }
      \label{tab:main-results}
    \end{table}

\subsection{Evaluation Protocol}
\label{sec:eval-protocol}
For each model, we provide a mature target topic $M$ and the 2019--2023 time window $t$, and ask the model to predict the weak signals that emerged during that period. The prediction prompt templates, which ask every system for at least ten weak signals ordered from most to least confident, are shown in are shown in \autoref{fig:pred-problem-prompt} and \autoref{fig:pred-solution-prompt}. Let the predicted weak-signal set be $s=\{s_1,\dots,s_m\}$ and the human-validated reference set be $g=\{g_1,\dots,g_n\}$. We report F1 under two LLM-judge settings with a strict matching criterion: two signals match only if they denote the same specific research topic, differing at most in wording; related, adjacent, or same-broad-area topics do not count. In \textbf{set-level LLM evaluation}, the judge sees $s$ and $g$ jointly and estimates set-level precision and recall, measuring alignment of the two sets as wholes. In \textbf{signal-level LLM evaluation}, the judge decides for each predicted signal whether it matches any reference signal, and symmetrically for each reference signal whether any prediction covers it, yielding a finer-grained measure of coverage. We also report \textbf{Coverage@K}, which measures the fraction of validated reference signals that are covered by the model's top-$K$ predictions. Specifically, for each reference signal, the judge determines whether it is matched by at least one of the top-$K$ predictions, and Coverage@K is the fraction of reference signals judged to be covered. The predictions are ordered from most to least confident, and we set $K=10$. Of the 50 topic--direction pairs formed by the \ntopic mature target topics and the two signal spaces, both settings score the 30 that carry at least one validated reference signal; the remaining 20 pairs have an empty reference set and are excluded because F1 is undefined when $g=\emptyset$, 14 of them belonging to the seven topics with no validated signal at all (\S\ref{app:absent}) and six being single-direction gaps within signal-bearing topics. All judgments use Claude-Opus-4.8, drawn from a model family independent of both the benchmark constructor and the evaluated systems; the judging prompts are shown in \autoref{fig:set-judge-prompt} and \autoref{fig:signal-judge-prompt}, and each setting is run 3 times with scores averaged. Embedding-similarity variants of these metrics reward shared domain vocabulary rather than same-topic identity, so we defer them to Appendix~\ref{app:bertscore}.  

To independently validate the automatic evaluation, two annotators assess every system’s predictions across all 30 topic-direction pairs with
validated weak signals using the same matching criterion as in the signal-level LLM evaluation. We report a Cohen’s $\kappa$ of 0.81, indicating substantial agreement between the two annotators. Human F1 in \autoref{tab:main-results} is aggregated as in the LLM-judge
columns: F1 per pair, then averaged across pairs.

\section{Experiments}

We use \ours to evaluate frontier LLMs, RAG systems, and DeepResearch agents along three axes: overall accuracy over the 30 topic-direction pairs with validated weak signals (\S\ref{sec:main-results}), the failure modes that arise when predictions miss the reference set (\S\ref{sec:error-analysis}), and analyses of the retrieval and web-search budgets (\S\ref{sec:analysis}).

\subsection{Main Results}
\label{sec:main-results}

\autoref{tab:main-results} shows that \ours is challenging even for strong LLM and DeepResearch systems. Under the LLM judge, set-level F1 ranges from 4.4\% to 10.1\% and signal-level F1 from 3.0\% to 7.1\%, with the human-judged F1 showing a similar range. Coverage@10 ranges from 8.4\% to 18.5\%, substantially higher than the corresponding signal-level F1 scores.

\paragraph{Problem signals are easier to recover.} 
According to the \autoref{tab:main-results}, weak-signal discovery is substantially harder in the solution space than in the problem space. All seven systems score higher on Problem than Solution under both set-level, signal-level F1 and Coverage@$10$. 

\paragraph{No simple capability hierarchy.}
\autoref{tab:main-results} shows that weak-signal discovery does not follow a clear model-capability hierarchy. DeepSeek-R1-0528 is consistently the strongest system across the LLM-judge metrics, but other frontier LLMs, retrieval-augmented models, and DeepResearch agents are interleaved rather than cleanly ordered by system type.

\paragraph{LLM judgments agree with human evaluation.}
The human largely agrees with the primary LLM judge: DeepSeek-R1-0528 and Qwen3-30B (RAG) rank first and second under both evaluations, while the remaining systems show broadly similar relative performance. This supports the reliability of the LLM-based evaluation.
 
\subsection{Error Analysis and Case Study}
\label{sec:error-analysis}

 We analyze GPT-5.4's predictions on all 30 topic--direction pairs with validated weak signals and identify four main error types:

  \noindent\textbf{(1) Topic drift.} Topic drift refers to cases where the prediction remains relevant to the target topic but shifts to an adjacent subproblem, meaning a related but different research question within the same literature that is not among the validated precursors. Although such predictions share vocabulary and context with the target topic, they fail to match the reference signals under the strict criterion of \S\ref{sec:eval-protocol}. 

  \noindent\textbf{(2) Granularity mismatch.} Granularity mismatch refers to cases where the prediction is plausible but at the wrong
  level of abstraction relative to the benchmark.

  \noindent\textbf{(3) Lexical near-miss.} Lexical near-miss refers to cases where the prediction is topically close to the
  benchmark but semantically different. 

  \noindent\textbf{(4) Coverage failure.} Coverage failure refers to cases where the model recovers one slice of the benchmark but
  misses other central dimensions. 

  The central challenge is thus \emph{semantic alignment with the intended precursor structure}: matching the right problem framing, abstraction level, and coverage. Appendix~\ref{app:error-cases} provides full examples of each error type, including the reference signals, model predictions, and their confidence ranks.

\begin{table}[htbp]
      \centering
      \small
      \setlength{\tabcolsep}{5pt}
      \renewcommand{\arraystretch}{1.15}
      \begin{tabular}{l l ccc}
      \toprule
      \textbf{Model} & \textbf{Depth} & \textbf{Set F1} & \textbf{Signal F1} & \textbf{Cov@10} \\
      \midrule
      \multirow{3}{*}{Qwen3-8B}
        & $k{=}10$ & 2.7 & 2.5 & 9.3 \\
        & $k{=}30$ & 4.0 & 3.1 & 9.8 \\
        & $k{=}50$ & \textbf{4.4} & \textbf{4.0} & \textbf{12.7} \\
      \cmidrule(lr){1-5}
      \multirow{3}{*}{Qwen3-30B}
        & $k{=}10$ & 5.2 & 4.1 & \textbf{10.4} \\
        & $k{=}30$ & 3.5 & 3.5 & 9.4 \\
        & $k{=}50$ & \textbf{6.6} & \textbf{5.0} & 9.8 \\
      \bottomrule
      \end{tabular}
      \caption{Retrieval-budget ablation (set-level / signal-level LLM-judge F1
  and Coverage@$10$, Claude-Opus-4.8, 3 runs; percentages). The $k{=}50$
  rows match the Qwen3~(RAG) rows of \autoref{tab:main-results}; best per model
  and metric in bold.}
      \label{tab:retrieval-budget}
    \end{table}

\subsection{Analysis}
\label{sec:analysis}
We conduct two analyses: examining the validity of unmatched predictions and probing evidence budgets in RAG and DeepResearch.

\paragraph{Unmatched predictions are often defensible.}
To assess how many unmatched predictions may represent valid weak signals absent from the reference set, we sampled 100 predictions from the 589 deduplicated unmatched predictions across all seven systems, stratified by system and signal direction. Two annotators independently judged whether each prediction was a defensible 2019--2023 weak signal for the given topic and direction, without access to the reference set. The two human annotators judged 57\% and 49\% of the predictions as plausible, respectively. This suggests that a substantial fraction of predictions treated as false positives by the reference-based evaluation may instead correspond to plausible weak signals that are absent from the reference set, indicating that the reference-based F1 may underestimate system performance.

\paragraph{Retrieval budget in RAG.}
We re-generate RAG predictions at retrieval depths $k\in\{10,30,50\}$ and evaluate them with the same LLM judge (\autoref{tab:retrieval-budget}). Both models achieve their best F1 at $k{=}50$, while Qwen3-8B improves consistently with increasing retrieval depth. Coverage@$10$ also increases for Qwen3-8B (9.3 to 12.7), whereas Qwen3-30B shows little benefit beyond $k{=}10$. These results suggest that retrieving more documents generally improves RAG performance, with a particularly strong effect on the smaller model.

\paragraph{Web search rounds for DeepResearch.}
We evaluate Tongyi-DR-30B-A3B with search-round budgets $b\in\{0,3,8, \infty\}$ (\autoref{tab:search-rounds}). Performance improves from no search to $b{=}8$ across the evaluated metrics, but declines under the unlimited policy. Coverage@$10$ increases from 6.6 at $b{=}0$ to 18.2 at $b{=}8$, before dropping to 10.4 with unlimited search. Thus, additional search rounds improve performance up to a point, while removing the search budget does not provide further gains.
\begin{table}[htbp]
      \centering
      \small
      \setlength{\tabcolsep}{6pt}
      \renewcommand{\arraystretch}{1.15}
      \begin{tabular}{l ccc}
      \toprule
      \textbf{Search budget} & \textbf{Set F1} & \textbf{Signal F1} & \textbf{Cov@10} \\
      \midrule
      $b{=}0$      & 4.9 & 2.7 & 6.6 \\
      $b{=}3$      & 5.7 & 2.8 & 16.6 \\
      $b{=}8$      & \textbf{7.9} & \textbf{4.4} & \textbf{18.2} \\
      $b{=}\infty$ & 5.6 & 4.3 & 10.4 \\
      \bottomrule
      \end{tabular}
      \caption{Web-search-round ablation for Tongyi-DR-30B-A3B, where $b$ is
  the number of search rounds the agent may issue; $b{=}\infty$ is the default
  policy of the main results, so its row equals the Tongyi row of
  \autoref{tab:main-results}. Set-level / signal-level LLM-judge F1
  and Coverage@$10$ (Claude-Opus-4.8, 3 runs; percentages); best per metric
  in bold.}
      \label{tab:search-rounds}
    \end{table}

\section{Conclusion}

We introduced \ours to test a notion of scientific foresight, centered on identifying the specific precursors that later mattered and recognizing when none exists. Our results expose less a knowledge gap than a discrimination gap: systems surface content in the right neighborhood but struggle with abstraction level, distinguishing genuine precursors from adjacent look-alikes, and covering a full direction. These failures persist with additional retrieval and search, suggesting that the bottleneck lies in evidence abstraction rather than access to evidence. The same conclusion holds under the more lenient Coverage@$K$: even without penalizing unmatched predictions, the best system recovers under a fifth of the reference signals. Scientific foresight therefore requires precise precursor identification beyond simply generating relevant scientific content. Progress on \ours will depend less on model scale or evidence budget than on abstraction-level control. More broadly, backward reconstruction offers a reusable recipe for foresight benchmarks with verifiable ground truth, applicable wherever a field's later record can adjudicate what was once a weak signal.

\section{Limitations}
\label{sec:limitations}

\ours's \ntopic mature target topics are all drawn from the Artificial Intelligence and Machine Learning domain of the 2024 JRC weak-signal report (\S\ref{sec:direction_focused}), providing a controlled but bounded view of established AI/ML research themes, so conclusions drawn from \ours should be read as evidence about how systems recover precursors of well-documented AI/ML topics rather than a measure of free-form scientific foresight, and extending the topic pool to other scientific domains as well as to longer-tail or pre-paradigmatic areas is a promising future direction.
Candidate precursors are mined from paper abstracts with a GPT-series LLM and then filtered by their 2019--2024 corpus frequency before expert validation, so the pipeline could inherit extraction- or labeling-specific phrasing biases from that model; we mitigate this by requiring every retained signal to be grounded in verifiable Semantic Scholar frequency and citation evidence rather than in model output alone, and by judging predictions with an independent Claude-Opus-4.8 evaluator drawn from a different model family than both the constructor and the evaluated systems (\S\ref{sec:direction_focused}, \S\ref{sec:eval-protocol}). A direct sensitivity study with alternative extraction models is a natural next step.
The end-of-2023 evidence cutoff is enforced by hard filtering of retrieved papers for the RAG and DeepResearch systems, but for parametric LLMs it can only be requested in the prompt (\S\ref{sec:evaluated-systems}). Their pretraining corpora extend past 2023, so we cannot rule out that a frontier model recalls how a target topic actually matured, and the foresight setting is therefore simulated rather than guaranteed for these systems. Any such leakage would inflate rather than depress the reported scores, so it does not weaken our central finding that every system scores at a low level; it does mean the parametric numbers should be read as an optimistic bound, and a strict test would require models whose pretraining cutoff precedes the prediction cutoff.
Following the practice of large-scale benchmark efforts~\citep{hendrycks2021measuringmassivemultitasklanguage, srivastava2023imitationgamequantifyingextrapolating, rein2023gpqagraduatelevelgoogleproofqa}, all validators are co-authors, and we reduce single-reviewer bias by running independent reviews before adjudication (\S\ref{sec:direction_focused}); recruiting a broader pool of external domain experts is a useful future extension.

\section*{Acknowledgments}
We thank TCS Research and the Yale NLP Lab for their support and helpful feedback.

\bibliographystyle{unsrtnat}
\bibliography{custom}

\appendix
\section{Appendix}
\subsection{Expert Information}
In this section, we provide an overview of the experts that contribute to the weak-signal validation and the human evaluation of model predictions. All experts are authors of the paper, so we do not provide monetary compensation to the experts.
\begin{table}[htbp]
\centering
\footnotesize
\renewcommand{\arraystretch}{1.15}
\begin{tabularx}{\textwidth}{@{}lllX@{}}
\toprule
\textbf{Expert} & \textbf{Level} & \textbf{Domain} & \textbf{Contribution} \\
\midrule
Expert 1 & Postdoctoral researcher & AI \& ML & Weak-signal validation \\
Expert 2 & 4th-year PhD student    & AI \& ML & Weak-signal validation \\
Expert 3 & 2nd-year PhD student    & AI \& ML & Human evaluation of model predictions \\
Expert 4 & 2nd-year MS student  & AI \& ML & Human evaluation of model predictions \\
\bottomrule
\end{tabularx}
\caption{Experts involved in \ours. Experts 1--2 independently validate the gate-passing
weak-signal candidates (\S\ref{sec:direction_focused}); Experts 3--4 independently annotate
all model predictions in the human evaluation (\autoref{tab:main-results}). All \ntopic{}
target topics fall in the Artificial Intelligence and Machine Learning domain, and all
experts are AI/ML researchers among the authors; the two weak-signal validators
(Experts 1--2) are the senior members of this group (\S\ref{sec:direction_focused}).}
\label{tab:expert_validation}
\end{table}

\subsection{A Running Example Through the Pipeline}
\label{app:running-example}
As a running example, take the mature target topic \emph{large language models}. Corpus construction retrieves $22.8$K papers for the topic and its paraphrases over 2019--2024 (\autoref{tab:domain_summary}). Candidate discovery mines solution-space labels from the 2019--2023 abstracts, among them \emph{retrieval-augmented language models}, and consolidation merges its surface variants into a single cluster. Frequency computation then yields the trajectory plotted in \autoref{fig:absent-traj}: normalized to its own early peak, the candidate stands at $0.03$ in 2019 and rises through $0.17$, $0.25$, and $0.45$ to $1.00$ in 2023, then reaches $2.95$ in 2024, measured through citations to its early papers. All four gates pass, since the early trace is faint, gap-free, and exponentially rising, and the 2024 lift of $2.95$ clears $\lambda=1.2$; expert validation retains it, and it becomes one of the topic's eleven validated signals. \autoref{fig:teaser} shows GPT-5.4 recovering exactly this signal, phrased as \emph{retrieval-augmented generation}, while missing others.

\subsection{The Growth Score of Gate $g_1$}
\label{app:score}
Gate $g_1$ of \S\ref{sec:direction_focused} tests $\mathrm{score}(c)>0$; we define that score here. It is evaluated once per candidate onset year $\tau\in\{2019,\dots,2022\}$, and the candidate's score and onset year are the maximum and the maximizer over $\tau$, ties broken toward the earliest onset.

Fix an onset $\tau$ and write $v^\tau=(f_\tau,\dots,f_{2023})$ for the frequency vector over the onset window, dropping the arguments of $f_y(c;M)$. Let $b_\tau$ and $R^2_\tau$ be the slope and the coefficient of determination of the least-squares fit of $\log(f_y+\epsilon)$ against $y-\tau$ over $v^\tau$, so that $b_\tau$ is the fitted exponential growth rate and $R^2_\tau$ measures how well a pure exponential describes the trajectory. Let $f^{+}_{\tau}$ and $m_\tau$ be the first nonzero entry and the number of nonzero entries of $v^\tau$, let $F_\tau=\max_{\tau\le y\le 2023} f_y$ and $F_{<\tau}=\max_{y<\tau} f_y$ with $F_{<2019}=0$, and set
\begin{align*}
\gamma_\tau &= \tfrac{f_{2023}+\epsilon}{f^{+}_{\tau}+\epsilon}, &
\pi_\tau &= \textstyle\frac{1}{2023-\tau}\sum_{y=\tau}^{2022}\mathbf{1}[f_{y+1}>f_y], &
\eta_\tau &= \min\!\big(1,\tfrac{f_{2023}+\epsilon}{F_{\tau}+\epsilon}\big),\\
\theta_\tau &= \min\!\big(1,\tfrac{f_{2023}+\epsilon}{f_{2022}+\epsilon}\big), &
\nu_\tau &= \min\!\big(1,\tfrac{m_\tau}{3}\big)^{2}, &
\omega_\tau &= \min\!\big(1,\tfrac{f_{\tau}+\epsilon}{F_{<\tau}+\epsilon}\big).
\end{align*}
These factors measure, in order, the total growth over the window, the fraction of years in which the frequency increases, how close the final year is to the window peak, how close it is to the preceding year, whether the candidate is supported in at least three distinct years, and how faint the candidate is at onset relative to anything before it. The onset score is
\[
s^d_\tau(c)=
\begin{cases}
0, & b_\tau\le 0,\\[2pt]
(R^2_\tau)^{2}\,\log^{+}\!\gamma_\tau\;\pi_\tau\,\eta_\tau\,\omega_\tau,
& d=\mathrm{prob},\\[2pt]
(R^2_\tau)^{3}\,\log^{+}\!\gamma_\tau\;\pi_\tau\,\eta_\tau\,\theta_\tau\,\nu_\tau\,\omega_\tau^{2},
& d=\mathrm{sol},
\end{cases}
\]
where $\log^{+}\!\gamma=\log\gamma$ for $\gamma>1$ and $0$ otherwise, and
\[
\mathrm{score}(c)=\max_{\tau} s^d_\tau(c),
\qquad
\tau_c=\arg\max_{\tau} s^d_\tau(c).
\]
The solution-space form is the stricter of the two: it additionally requires support in several years and penalizes a final-year dip and pre-onset visibility more heavily, matching the fact that a method is adopted gradually whereas a problem can be posed in a single influential paper.

Two properties of this construction matter for how $g_1$ behaves. First, every factor is non-negative, so $\mathrm{score}(c)>0$ holds exactly when all of them are positive: the fitted growth rate must satisfy $b_\tau>0$, the trajectory must grow overall ($\gamma_\tau>1$) and rise in at least one year ($\pi_\tau>0$), the exponential fit must be non-degenerate ($R^2_\tau>0$), and the candidate must be present in 2023. Flat, declining, and single-point traces therefore score zero at every onset. Second, because the gate tests only positivity, the exponents on $R^2_\tau$ and $\omega_\tau$ affect only how surviving candidates are ranked, never which ones pass.

\subsection{Analysis of Survey Paper Exclusion Strategy}
\label{app:surveys}
The frequency $f_y(c;M)$ of \S\ref{sec:direction_focused} counts non-survey papers only, so surveys are removed at the counting stage rather than before candidate extraction. To examine this choice, we rerun the full pipeline over all \ntopic{} topics with survey-titled papers excluded before candidate extraction.

Surveys do not contribute frequency evidence under either setting: $n_y(c)$ and $N_y(M)$ count non-survey papers only. For the $99.68\%$ of candidates whose cluster composition is unchanged, the entire 2019--2023 trajectory is identical across the two runs. However, surveys can provide additional terminology during candidate discovery: survey abstracts often summarize multiple research directions explicitly, supplying candidate labels that help clusters match additional \emph{non-survey} papers. Excluding surveys earlier removes vocabulary from $453$ clusters, and $94$ of these lose non-survey supporting papers as a result, whereas unchanged clusters lose none.

Under our benchmark, the earlier exclusion strategy recovers fewer weak signals ($117$ vs. $125$) and fewer human-validated signals ($61$ vs. $66$). Survey-titled papers constitute only $1.36\%$ of the retrieved corpus, so the overall effect is limited; nevertheless, the counting-stage exclusion separates the two roles of surveys: they assist candidate discovery while not contributing to the evidence used for validation.

\subsection{Empirical Validation of the Growth Criterion}
\label{app:growth-validation}
Unlike a purely qualitative judgment, \ours's growth requirement is enforced directly during construction rather than checked post hoc. A candidate is retained only if its 2024 frequency exceeds its peak 2019--2023 frequency by a factor of $\lambda=1.2$ (gate $g_2$ in \S\ref{sec:direction_focused}), so every validated weak signal is, by construction, more frequent in 2024 than at any point during its early emergence window. Expert validation then confirms that this measured rise reflects a genuine research trajectory rather than a measurement artifact. Because the criterion is embedded in selection, all \nweak{} validated signals satisfy the growth condition.

\subsection{Two Cases That Warrant Explanation}
\label{app:two-cases}

A weak signal is defined relative to a mature target topic's own corpus and to a designated signal space (\S\ref{sec:backtrend-task}). Under this definition, two cases in the final \nweak{}-signal set may look surprising at first, and we explain them here.

\paragraph{A mature target topic can itself be a weak signal of another topic.}
Machine unlearning is one of our \ntopic{} mature target topics, yet it is also a validated problem-space weak signal of privacy-preserving machine learning. The underlying reason is that the JRC mature topics are not all at the same level of granularity: the report mixes broad research areas with much narrower directions, so a narrower topic can sit inside a broader one. Our definition handles this cleanly, because maturity and weak-signal frequency are measured in different corpora. A topic's maturity is established within its own corpus, whereas its weak-signal frequency is measured within the target topic's corpus (\S\ref{sec:direction_focused}). As a standalone field machine unlearning is mature, but within the broader privacy-preserving-ML literature it was a low-visibility precursor over 2019--2023 whose frequency rose sharply into 2024, which is exactly the kind of signal the benchmark is meant to recover.

\paragraph{A single direction can occupy both signal spaces of one topic.}
Within machine unlearning, federated unlearning appears as a weak signal in both the problem space and the solution space, the only such case among the \nweak{} signals. The two occurrences are reasonable because they carry a different focus. As a problem-space signal, federated unlearning is the open question of how to guarantee that a client's contribution can be removed from a trained federated model and how to verify that the removal took place. As a solution-space signal, it is a concrete method that erases a client's data without retraining the model from scratch. To confirm that these are genuinely two signals rather than one label counted twice, we inspected their supporting paper pools and found them disjoint, so each occurrence rests on its own distinct evidence, and expert validation retained both.

\subsection{Why Seven Mature Topics Carry No Certifiable Weak Signal}
\label{app:absent}

Of the \ntopic{} mature target topics, \nweak{} validated weak signals are distributed over 18 topics, while seven carry none (\autoref{tab:domain_summary}). All \ntopic{} target topics, absent and present alike, were fixed in advance by the same external JRC report (\S\ref{sec:direction_focused}); which of them turn out empty is determined by the construction alone. Nor is an empty set anomalous under our definition. A weak signal is a discrete research direction that (i) was of low visibility early, (ii) grew exponentially within the 2019--2023 window, and (iii) subsequently rose in the mature topic's 2024 corpus (\S\ref{sec:backtrend-task}): clauses (i)--(ii) require the emergence to be measurable in the historical record, so that a foresight system at the end-2023 cutoff could in principle detect it, and clause (iii) verifies that the emergence was real. A topic can therefore mature along paths that leave no such trace in its own corpus: by aggregating sub-areas that were already visible, so nothing satisfies (i); by drawing its 2024 growth from work outside the topic's corpus boundary, so nothing satisfies (ii)--(iii) together; or by advancing on a corpus too small to measure any yearly frequency above noise, so nothing satisfies (ii) credibly. This subsection shows that the seven empty sets are of exactly this structural kind by ruling out each stage of the construction in order: the paper pools, the candidate clustering, the four gates, and expert validation. Five of the seven nulls are decided by the same external $g_2$ frequency gate that certifies the other 18 topics and two by the same expert criteria that curate every retained signal; for none does the pipeline find a candidate satisfying the definition with a measurable 2024 rise, so we retain them with documented empty signal sets.

\paragraph{Searching harder changes nothing.}
The most natural objection is that these topics were simply not searched thoroughly enough. We therefore re-ran retrieval for every topic with substantially expanded paraphrase-query sets, after a semantic-noise cleanup of overly broad queries, and re-ran the \emph{entire} pipeline on the enlarged pools: candidate extraction, clustering, frequency computation, and gating are all recomputed from scratch. The pools of four of the five gate-failing topics grew by $+24\%$ to $+93\%$, and those of the two expert-null topics by $+37\%$ and $+232\%$, yet none produced a validated weak signal; the fifth gate-failing topic, \emph{Federated RL}, grew by under $1\%$, so its corpus is intrinsically small; and a twelve-variant leave-one-in paraphrase sweep for the most fragile case, \emph{Epistemic AI}, produced zero signals throughout. Because the deciding lift is a ratio of within-corpus frequencies, only a shift in the topical \emph{composition} of the yearly corpora could move it; the expansion shifts composition substantially and moved none of these topics across the gate.

\paragraph{Re-merging clusters at any granularity changes nothing.}
A subtler objection holds that the cosine-$0.85$ consolidation (\S\ref{sec:direction_focused}) shattered one genuine direction into near-duplicate labels, each individually too sparse to certify. We therefore re-merged the final clusters of every absent topic by single-linkage in the original embedding space at seven coarser thresholds, from $0.80$ down to $0.50$, recomputing each merged union's trajectory exactly, as deduplicated unions of supporting and 2024 citing papers under the pipeline's counting rules. Single-linkage merging and a permissive reading of the four gates both favor the objection, yet of the $5{,}411$ merged unions produced, only eleven union--threshold instances pass, all in the narrow band $0.80$--$0.75$ and none below: coarser merging adds early-window mass faster than 2024 mass, so the low-visibility requirement collapses first. Each of the eleven reduces, on inspection, to a case adjudicated below. In nine, a single member carries $86$--$100\%$ of the union's 2024 citations and every other member is a single-year blip contributing $0$--$2$ citations; the sharpest, a few-/zero-shot multimodal learning union with trace $0,0,0,1,5$, draws \emph{all} 29 of its 2024 citations from one member's single 2022 paper, Flamingo~\citep{alayrac2022flamingo}. Such a union staples four unrelated blips onto the citation halo of an already-famous paper, the late-arriving pattern analyzed below. The remaining two pool the \emph{multimodal reasoning} family and \emph{Federated DL}'s drift-and-heterogeneity vocabulary, and inherit those cases' defects: the merged reasoning trace sits at $48\%$ of its own peak frequency already in 2019, and the merged drift union's frequency \emph{falls} into the cutoff, $0.0217\!\to\!0.0180$. A genuinely fragmented emergence would spread its 2024 citations across the reunited fragments and rise as a whole; no union at any granularity does. The sparsity is a property of these literatures themselves.

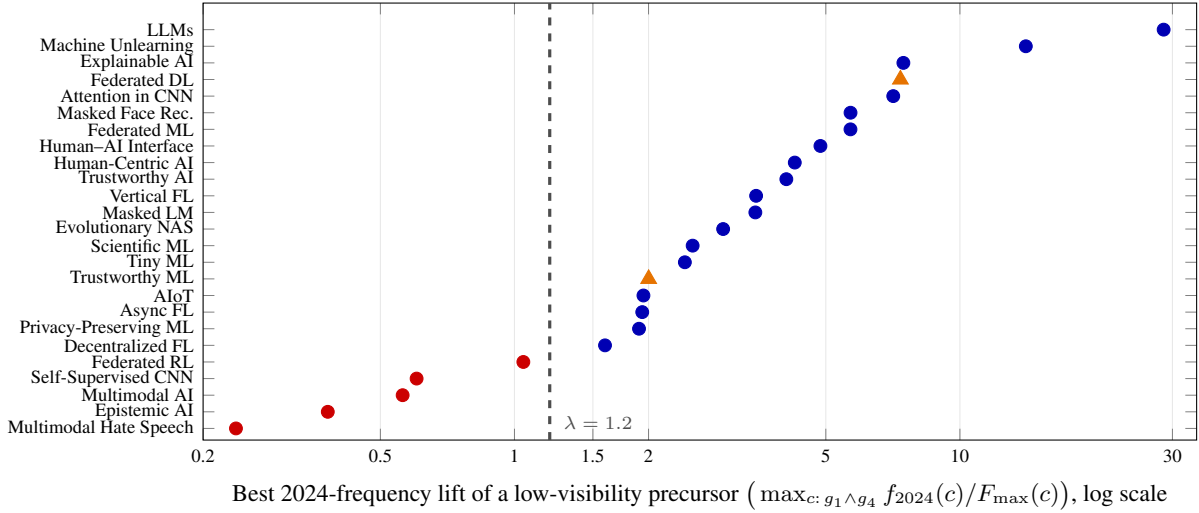
\begin{figure}[tbp]
\centering
\begin{tikzpicture}
\begin{axis}[
  width=0.92\textwidth, height=0.46\textwidth,
  xmode=log, log basis x=10,
  xmin=0.2, xmax=34, ymin=0.3, ymax=26.6,
  ytick={1,2,3,4,5,6,7,8,9,10,11,12,13,14,15,16,17,18,19,20,21,22,23,24,25}, yticklabels={Multimodal Hate Speech,Epistemic AI,Multimodal AI,Self-Supervised CNN,Federated RL,Decentralized FL,Privacy-Preserving ML,Async FL,AIoT,Trustworthy ML,Tiny ML,Scientific ML,Evolutionary NAS,Masked LM,Vertical FL,Trustworthy AI,Human-Centric AI,Human--AI Interface,Federated ML,Masked Face Rec.,Attention in CNN,Federated DL,Explainable AI,Machine Unlearning,LLMs},
  yticklabel style={font=\scriptsize},
  xlabel={\small Best 2024-frequency lift of a low-visibility precursor $\big(\max_{c:\,g_1\wedge g_4} f_{2024}(c)/F_{\max}(c)\big)$, log scale},
  xtick={0.2,0.5,1,1.5,2,5,10,30}, xticklabels={0.2,0.5,1,1.5,2,5,10,30},
  xticklabel style={font=\scriptsize},
  xmajorgrids, grid style={gray!18}, clip=false,
]
\addplot[domain=0.3:26.6, samples=2, dashed, very thick, black!70] ({1.2},x);
\node[anchor=south west, font=\scriptsize\bfseries, black!70] at (axis cs:1.23,0.4) {$\lambda=1.2$};
\addplot[only marks, mark=*, mark size=2.4pt, blue!70!black] coordinates {(1.597,6) (1.904,7) (1.936,8) (1.948,9) (2.413,11) (2.512,12) (2.941,13) (3.474,14) (3.488,15) (4.077,16) (4.259,17) (4.863,18) (5.680,19) (5.682,20) (7.086,21) (7.467,23) (14.060,24) (28.669,25)};
\addplot[only marks, mark=*, mark size=2.4pt, red!80!black] coordinates {(0.237,1) (0.381,2) (0.561,3) (0.603,4) (1.047,5)};
\addplot[only marks, mark=triangle*, mark size=3.4pt, orange!90!black] coordinates {(2.001,10) (7.353,22)};
\end{axis}
\end{tikzpicture}
\caption{\textbf{Why seven mature topics carry no certifiable weak signal: absence is decided by the same 2024-frequency gate that certifies the present topics.} For each of the \ntopic{} target topics we plot the \emph{largest} 2024-frequency lift achieved by any candidate passing the low-visibility and exponential-growth gates ($g_1\wedge g_4$); a validated weak signal requires this lift to clear gate $g_2$ ($\lambda=1.2$, dashed line). \textcolor{blue!70!black}{$\bullet$~All 18 signal-bearing topics} clear the line ($1.60$--$28.7$). \textcolor{red!80!black}{$\bullet$~Five absent topics} fall below it (all $\le 1.05$, even though a per-topic maximum over noisy candidate ratios is biased upward): point estimates decline for four and are flat for Federated RL ($1.05$, CI $[0.38,2.92]$); \S\ref{app:absent} additionally audits every candidate above the line that the other gates excluded. \textcolor{orange!90!black}{$\blacktriangle$~Two absent topics} clear the line, but expert validation removes every surviving candidate: none of their year-by-year traces is a genuine exponential rise (\S\ref{app:absent}). Topic names are abbreviated on the axis; \autoref{tab:domain_summary} lists them in full.}
\label{fig:absent-lift}
\end{figure}

\paragraph{The gates decide the five automatic nulls and separate the two groups cleanly.}
With the pools and the clustering ruled out, the decision rests with the four gates of \S\ref{sec:direction_focused}, which operationalize the three defining clauses: $g_1$ tests for an exponentially rising early trajectory, $g_4$ for low pre-onset visibility, $g_3$ for persistent growth, and the decisive $g_2$ requires a candidate's 2024 frequency to exceed its own early peak by a factor $\lambda=1.2$. Gate $g_2$ is the only \emph{external} check, grounded in the candidate's independently measured 2024 frequency, and it is the one that decides the five automatic nulls. Across the \ntopic{} topics the pipeline mines $133{,}426$ candidate topics; for each topic we take its \emph{best} low-visibility precursor, the candidate maximizing the 2024-frequency lift $f_{2024}(c)/F_{\max}(c)$ among those already passing $g_1$ and $g_4$, and compare it against the $g_2$ threshold (\autoref{fig:absent-lift}). All 18 signal-bearing topics clear the threshold with best lifts of $1.60$--$28.7$; each yields $2$--$24$ gate-passing candidates, from which expert validation retains the final \nweak{} signals. Five of the seven absent topics fall \emph{below} the threshold, all at best lift $\le 1.05$. This statistic is generous to them, since a per-topic maximum over many noisy per-candidate ratios is biased \emph{upward} by selection. The instrument itself holds up. Renaming cannot produce a null: the 2024 frequency is measured through citations to the candidate's early papers and follows the line of work however it is later phrased (\S\ref{sec:direction_focused}); the one thing no within-corpus measure can follow is work migrating across the \emph{target topic's} corpus boundary. This is by design, since the task asks for precursors \emph{of} $M$ certified in $M$'s own literature. The asymmetry of the measurement, in-corpus matching for 2019--2023 and citation-based for 2024, cannot manufacture the shortfalls either: within these very corpora the same instrument registers large 2024 rises for other candidates, lift $10.1$ with CI $[1.3,75.5]$ in \emph{Multimodal AI} and lift $10.0$ in \emph{Federated DL}, both from the riser audit below. What the gate actually observes is decline: the point estimates fall for four of the five topics and are flat for the fifth, \emph{Federated RL}, and where the counts are large enough the decline is statistically certified below. Rejecting a topic whose best precursor shows no 2024 rise is the same gate that guarantees the growth property for the present topics (\autoref{app:growth-validation}) behaving \emph{correctly} under the definition. One route remains open: a genuine riser could in principle have been discarded by the \emph{other} gates; the audit below closes it.

\paragraph{The gates discard no certifiable riser.}
The lift statistic above conditions on $g_1\wedge g_4$; could those gates themselves be hiding a genuine signal? We therefore audit, for the five gate-failing topics, \emph{every} mined candidate whose measured 2024 frequency clears $\lambda$ when all other gates are ignored: 66 candidates in total, a count that only coincidentally equals that of the validated signals. Each falls into one of three classes. \emph{(a) Single-year blips} (57 of 66): candidates whose entire 2019--2023 record is at most three papers, all in a single year. The sharpest example is \emph{domain-specific multimodal reasoning} (\emph{Multimodal AI}): one 2023 paper, 18 citing papers in 2024, lift $10.1$, nominally even a statistically significant rise ($95\%$ confidence interval $[1.3,75.5]$; Katz log-ratio interval for a ratio of two binomial proportions, here and throughout). It still certifies nothing about emergence: growth, exponential or otherwise, is not observable from a single point, so clause (ii) fails outright. Admitting such candidates would turn ground truth into a post-hoc lottery: $86$--$92\%$ of \emph{all} candidates mined for these five topics have single-year records ($3{,}943$ of \emph{Multimodal AI}'s $4{,}469$), only $0.5$--$6\%$ of those happen to rise in 2024, and at the end-2023 cutoff nothing in the record separates the eventual winners from the rest. No validated signal is of this form: all \nweak{} kept signals have at least two nonzero in-window years. \emph{(b) Sparse broken traces} (8 of 66): two or three nonzero years at the 1--2-paper level whose trace is gapped ($1,0,0,0,1$; $0,0,1,0,1$) or flat-to-falling in frequency, so no positive frequency-growth fit exists; these are exactly the shapes expert validation removes wherever they do pass the gates (next paragraph). \emph{(c) Already visible} (1 of 66): \emph{multimodal reasoning} (\emph{Multimodal AI}), at $61\%$ of its own peak frequency already in 2019, violating clause (i); its lift of $1.36$ moreover carries a CI of $[0.57,3.28]$, so even its rise is uncertifiable. Extending the audit to the two expert-null topics adds no new class: beyond the nine survivors reviewed next, their 84 unconditional risers decompose into 67 single-year blips, 12 gapped 1--2-paper traces, four candidates whose \emph{frequency} is flat or falling even as raw counts grow, and one $g_4$-rejected candidate, \emph{contrastive learning in federated learning} at lift $10.0$, which anticipates the natural experiment below: the same direction passes every gate in the sibling \emph{Federated ML} corpus and is kept there. Across all seven topics, the gates therefore discard nothing that our definition, or any definition requiring emergence to be measurable in the 2019--2023 record, would admit.

\paragraph{Expert validation decides the remaining two, on the grounds it applies everywhere.}
\emph{Trustworthy ML} and \emph{Federated DL} clear the gates with 6 and 3 automatic survivors, and expert validation removes every one. This stage applies the validation criteria of \S\ref{sec:direction_focused} everywhere; miscategorization and fixable evidence issues are handled by revision, so a survivor is \emph{removed} only on two grounds: (i) it does not name an AI/ML research direction, or (ii) its year-by-year trajectory lacks the low-then-exponentially-rising shape the definition requires (\S\ref{sec:backtrend-task}). Across the 18 signal-bearing topics these grounds keep 66 of 116 survivors; ground (i) accounts for a single removal, \emph{transparent face masks} under \emph{Masked Face Recognition}, which passes every statistical gate yet names a physical object; every other removal is ground (ii). Ground (ii) judges the full trace shape, and its footprint is auditable. The 59 removed survivors comprise the 50 from the signal-bearing topics plus the nine here; 44 of them, $75\%$, contain an interior zero year, versus 2 of the 66 kept signals, and the other 64 kept traces run gap-free from onset to 2023. All nine survivors of the two absent topics fall to ground (ii). Each \emph{Trustworthy ML} survivor rests on one or two papers per nonzero year with multi-year gaps, traces such as $1,0,0,0,1$ or $0,1,1,0,1$ over 2019--2023: on counts this small the log-space fit behind $g_1$ can turn positive, but the trace is visibly noise. In a corpus of $2{,}200$--$4{,}600$ papers per year over 2019--2023, that is itself the structural finding: no genuine precursor ever concentrates under the umbrella vocabulary. \emph{Federated DL}'s survivors fail identically: \emph{client drift} rises in raw counts but is flat in frequency, $0.0032\!\to\!0.0041\!\to\!0.0033$, while \emph{prototype-based FL} and \emph{batch normalization in FL} both trace $0,0,1,0,3$. \emph{Prototype-based FL} is instructive: the \emph{same} precursor rises monotonically in the broader \emph{Federated ML} corpus, tracing $0,0,0,1,3$ at lift $2.13$, and is kept \emph{there} as a validated signal; its certifiable rise simply lives in the sibling's corpus.

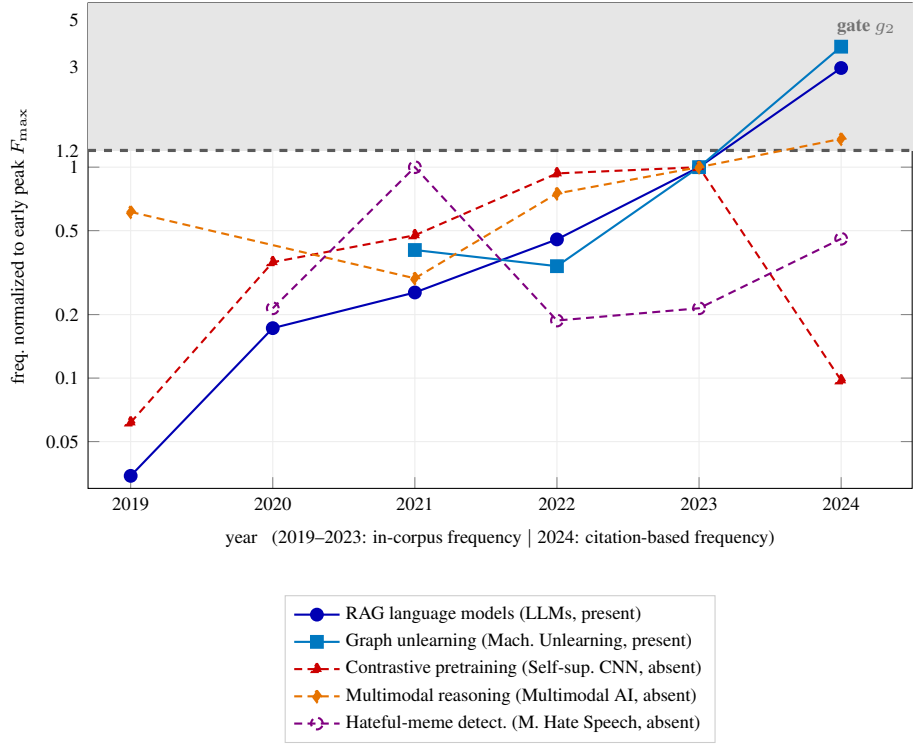
\begin{figure}[tbp]
\centering
\begin{tikzpicture}
\begin{axis}[
  width=0.78\linewidth, height=0.50\linewidth,
  ymode=log, log basis y=10,
  xmin=2018.7, xmax=2024.5, ymin=0.03, ymax=6,
  xtick={2019,2020,2021,2022,2023,2024},
  xticklabels={2019,2020,2021,2022,2023,2024}, xticklabel style={font=\scriptsize},
  yticklabel style={font=\scriptsize},
  ytick={0.05,0.1,0.2,0.5,1,1.2,3,5}, yticklabels={0.05,0.1,0.2,0.5,1,1.2,3,5},
  xlabel={\scriptsize year \; (2019--2023: in-corpus frequency $\vert$ 2024: citation-based frequency)},
  ylabel={\scriptsize freq.\ normalized to early peak $F_{\max}$},
  legend style={font=\scriptsize, at={(0.5,-0.22)}, anchor=north, draw=gray!40, fill=white, legend columns=1},
  legend cell align=left, grid=both, grid style={gray!14}, clip=false,
]
\fill[gray!20] (axis cs:2018.7,1.2) rectangle (axis cs:2024.5,6);
\addplot[dashed, very thick, black!65, domain=2018.7:2024.5, samples=2, forget plot]{1.2};
\node[anchor=north east, font=\scriptsize, black!55] at (axis cs:2024.45,5.6) {\textbf{gate} $g_2$};
\addplot[solid, thick, blue!70!black, mark=*, mark size=2.2pt] coordinates {(2019,0.0344) (2020,0.1725) (2021,0.2546) (2022,0.4545) (2023,1.0000) (2024,2.9515)};
\addlegendentry{\scriptsize RAG language models (LLMs, present)}
\addplot[solid, thick, cyan!55!blue, mark=square*, mark size=2.2pt] coordinates {(2021,0.4047) (2022,0.3393) (2023,1.0000) (2024,3.7254)};
\addlegendentry{\scriptsize Graph unlearning (Mach.\ Unlearning, present)}
\addplot[densely dashed, thick, red!80!black, mark=triangle*, mark size=2.2pt] coordinates {(2019,0.0616) (2020,0.3544) (2021,0.4753) (2022,0.9340) (2023,1.0000) (2024,0.0975)};
\addlegendentry{\scriptsize Contrastive pretraining (Self-sup.\ CNN, absent)}
\addplot[densely dashed, thick, orange!90!black, mark=diamond*, mark size=2.2pt] coordinates {(2019,0.6135) (2021,0.2971) (2022,0.7503) (2023,1.0000) (2024,1.3615)};
\addlegendentry{\scriptsize Multimodal reasoning (Multimodal AI, absent)}
\addplot[densely dashed, thick, violet, mark=o, mark size=2.2pt] coordinates {(2020,0.2143) (2021,1.0000) (2022,0.1875) (2023,0.2143) (2024,0.4576)};
\addlegendentry{\scriptsize Hateful-meme detect.\ (M.\ Hate Speech, absent)}
\end{axis}
\end{tikzpicture}
\caption{\textbf{The weak-signal signature and three absent-topic failure modes.} Each candidate's yearly frequency is normalized to its own 2019--2023 peak $F_{\max}$, so every trajectory reaches $1.0$ in its early window; the 2024 point is the citation-based 2024 frequency, and entering the shaded band ($>\!\lambda F_{\max}$, $\lambda=1.2$) is exactly the gate $g_2$. \textbf{Present} signals rise from low visibility and then leap into the band in 2024. The three \textbf{absent}-topic candidates each fail differently: \emph{contrastive pretraining} rose from low visibility to $11\%$ of the CNN-scoped 2023 corpus, then its frequency \emph{collapses} in 2024 as the method matured onto transformers (lift $0.10$, CI $[0.05,0.21]$: a statistically certified failure of the outcome condition); \emph{multimodal reasoning} was never low-visibility, starting at $61\%$ of its own peak, so $g_4$ rejects it, and its 2024 rise is itself uncertifiable (CI $[0.57,3.28]$); \emph{hateful-meme detection} spikes on a handful of papers in a tiny corpus and collapses, noise with no positive growth fit for $g_1$. None completes the low-visibility, in-window-rise, 2024-confirmation signature that defines a weak signal.}
\label{fig:absent-traj}
\end{figure}

\paragraph{Each absence traces to a measurable structural cause.}
\autoref{tab:absent-taxonomy} assigns each absent topic to its dominant cause, and \autoref{fig:absent-traj} shows the mechanism directly by plotting normalized frequency trajectories. (1) \emph{Umbrella or narrowly scoped topics whose gate-passing survivors are sparse-count artifacts}: \emph{Trustworthy ML} and \emph{Federated DL}, as established above. (2) \emph{Scope migration: topics whose growth crosses the corpus boundary.} \emph{Multimodal AI}'s vocabulary consolidated only recently: its corpus quadruples from 448 papers in 2019 to $1{,}924$ in 2023, and its 2024 growth is carried by directions with no in-corpus history. Of its 29 unconditional risers, 25 are the single-year blips of class (a), 19 of them entering only in 2023, yet drawing 3--29 citing 2024 papers each. Sub-areas such as visual language models and vision-language instruction tuning arrive in the corpus already formed, which is why its best low-visibility precursor reaches a lift of only $0.56$. \emph{Self-Supervised CNN} shows the outbound migration: its genuine early risers, contrastive and masked pretraining, matured and moved to transformer backbones, so within the CNN-scoped corpus their frequency collapses in 2024 (\autoref{fig:absent-traj}), and the counts are large enough to \emph{certify} the collapse. \emph{Contrastive pretraining} grows from 1 to 65 supporting papers over 2019--2023, reaching $11\%$ of the 2023 corpus, then falls to 7 of 664 in 2024, a lift of $0.10$ with CI $[0.05,0.21]$; \emph{masked image modeling} behaves identically at lift $0.09$, CI $[0.01,0.71]$. This topic \emph{had} real early risers; what it lacks is any precursor whose rise survives into its own 2024 corpus, the defining outcome condition (iii). A 2023-vintage forecaster would have named contrastive pretraining, and 2024 proved that call wrong; the empty set records precisely this. (3) \emph{Sparse or fragmented corpora}, small in \emph{every} year of 2019--2023, so no choice of onset year escapes the sparsity. \emph{Multimodal Hate Speech}, with 3--29 papers per year, is noise-dominated: its best low-visibility candidate rests on 2 papers at its peak against a single citing paper in 2024, a lift of $0.24$ with CI $[0.02,2.51]$; at counts this small no trajectory, rising or falling, can be certified under \emph{any} threshold, and that impossibility is itself the intrinsic property. \emph{Epistemic AI}, with 53--270 papers per year split across unrelated senses of ``epistemic'', still yields a statistically solid negative: its strongest candidate falls from $37/188$ source papers at its 2022 peak to $30/400$ in 2024, a lift of $0.38$ whose CI $[0.24,0.60]$ lies entirely below the gate. \emph{Federated RL}, with 69--381 papers per year, is the one borderline case: its corpus can neither certify a rise nor rule one out, as quantified below.

\begin{table}[htbp]
\centering\small
\setlength{\tabcolsep}{4pt}\renewcommand{\arraystretch}{1.15}
\begin{tabular}{l r c l p{6.6cm}}
\toprule
\textbf{Absent topic} & \textbf{\#\,cand.} & \textbf{best lift} & \textbf{Outcome} & \textbf{Dominant cause} \\
\midrule
Trustworthy ML & 16776 & 2.00 & 6 survivors, 0 kept & All 6 removed because their frequency trajectories are not exponential rises: each rests on 1--2 papers per nonzero year with multi-year gaps (e.g.\ $1,0,0,0,1$), visibly noise. Nothing concentrates under the umbrella vocabulary. \\
Federated DL & 1781 & 7.35 & 3 survivors, 0 kept & All 3 removed because their frequency trajectories are not exponential rises: erratic ($0,0,1,0,3$) or frequency-flat; the one genuine precursor (\emph{prototype-based FL}) rises monotonically, and is kept, in the broader \emph{Federated ML} corpus. \\
Multimodal AI & 4469 & 0.56 & $0$ pass $g_1\!\cdots g_4$ & Late-consolidating container: corpus quadruples 2019--2023, and its 2024 growth is carried by candidates with single-year in-corpus records (25 of its 29 risers, mostly 2023-only); the one already-visible riser (\emph{multimodal reasoning}) is rejected by $g_4$, and its rise is uncertifiable (CI $[0.57,3.28]$). \\
Self-Supervised CNN & 1960 & 0.60 & $0$ pass $g_1\!\cdots g_4$ & Scope migration: precursors matured onto transformer backbones; their in-corpus collapse is statistically certified (\emph{contrastive pretraining} $65\!\to\!7$ papers, lift $0.10$, CI $[0.05,0.21]$). \\
Federated RL & 539 & 1.05 & $0$ pass $g_1\!\cdots g_4$ & Sparse niche (69--381 papers/yr, 2019--23): paraphrase expansion adds $<\!1\%$ papers; best lift statistically indistinguishable from flat (CI $[0.38,2.92]$). \\
Epistemic AI & 948 & 0.38 & $0$ pass $g_1\!\cdots g_4$ & Sparse (53--270 papers/yr, 2019--23), fragmented across unrelated senses of ``epistemic''; strongest candidate's decline is statistically certified (lift $0.38$, CI $[0.24,0.60]$). \\
Multimodal Hate Speech & 112 & 0.24 & $0$ pass $g_1\!\cdots g_4$ & Extreme sparsity (3--29 papers/yr, 2019--23): best candidate's CI $[0.02,2.51]$ spans the gate: too few papers to certify any trajectory, under any threshold. \\
\bottomrule
\end{tabular}
\caption{The seven mature topics with no validated weak signal. \textbf{\#\,cand.}: candidate topics mined; \textbf{best lift}: largest 2024-frequency lift among low-visibility precursors ($g_1\!\wedge\! g_4$), which gate $g_2$ requires to reach $\lambda\!=\!1.2$; \textbf{Outcome}: result of the automatic gates and expert validation. Full analysis in \S\ref{app:absent}.}
\label{tab:absent-taxonomy}
\end{table}

\paragraph{Design choice: null sets over manufactured signals.}
We could have relaxed $\lambda$ or the onset gates until every topic yielded a candidate, but the riser audit above shows exactly what relaxation would admit: single-year blips, gapped one-paper traces, and already-visible risers, all false positives by definition, eroding the very property that distinguishes \ours from unvalidated trend lists. The choice $\lambda=1.2$ carries no responsibility for the null cases either: sweeping the threshold anywhere above $1.05$ leaves all seven topics empty. For the five gate-failing topics this is immediate, since their best eligible lift is $1.047$; for the two expert-null topics, every candidate that passes the gates at any $\lambda\ge 1.05$ is one of the nine survivors already reviewed and removed above. An absent topic first gains a candidate only when $\lambda$ is pushed below $1.05$: the sole such topic is \emph{Federated RL}, whose best candidate appears in $6/372$ papers at its 2023 peak and $9/533$ in 2024, a lift of $1.05$ with a $95\%$ confidence interval of $[0.38,2.92]$, statistically indistinguishable from flat on a $1{,}600$-paper corpus. We instead retain the seven topics as target topics with documented empty signal sets: this preserves benchmark integrity and supplies honest negative cases against which a well-calibrated foresight system should \emph{abstain}. No evaluated system currently does: prompted on these seven topics, every system of \S\ref{sec:evaluated-systems} still returns 3--13 predicted precursors per topic--direction pair, $6.1$ on average across all 98 system--pair combinations, and abstains on none, precisely the behavior these documented nulls make measurable. That seven of \ntopic{} externally chosen mature topics genuinely lack a certifiable precursor trace is itself a finding about how fields mature: not all growth is preceded by a detectable weak signal, and a benchmark unable to represent this outcome would presuppose exactly what foresight systems are meant to establish.

\subsection{Gate-Passing Candidates Removed by Expert Validation}
\label{app:removed}
The four automatic gates of \S\ref{sec:direction_focused} pass 125 candidates across the \ntopic{} mature target topics, of which expert validation retains \nweak{}. \autoref{tab:removed-candidates} names the 59 removals, grouped by target topic and signal space, so that the outcome of validation is auditable per topic rather than only in aggregate. Removals concentrate in the topics with the largest candidate pools, \emph{Large Language Models} alone accounting for 13, and they are dominated by trajectory shape rather than topical error: exactly one candidate is removed for not naming a research direction, \emph{transparent face masks} under \emph{Masked Face Recognition}, while every other removal is a trace that lacks the low-then-exponentially-rising shape the definition requires. That is the same ground on which \S\ref{app:absent} empties the signal sets of \emph{Trustworthy ML} and \emph{Federated DL}, so those two topics are not adjudicated by a stricter standard than the rest.

\begin{table}[htbp]
\centering
\footnotesize
\setlength{\tabcolsep}{4pt}
\renewcommand{\arraystretch}{1.15}
\begin{tabular}{@{}l c >{\raggedright\arraybackslash}p{0.60\linewidth}@{}}
\toprule
\textbf{Mature Target Topic} & \textbf{\#} & \textbf{Gate-passing candidates removed by expert validation} \\
\midrule
Artificial Intelligence of Things & 4 & \emph{(P)} cyber-physical production systems interoperability; heterogeneity in federated learning for IoT; intrusion detection in Internet of Vehicles. \emph{(S)} AI-assisted network slicing \\
\midrule
Attention Mechanisms in CNN & 2 & \emph{(P)} local feature modeling in vision transformers. \emph{(S)} deformable self-attention \\
\midrule
Evolutionary Neural Arch.\ Search & 2 & \emph{(P)} neural architecture size--performance tradeoffs. \emph{(S)} weight inheritance in evolutionary neural architecture search \\
\midrule
Explainable AI & 6 & \emph{(P)} XAI deployment challenges; actionability of AI explanations; explainable fake news detection; explanation effectiveness. \emph{(S)} explanation method benchmarking; layer-wise relevance propagation \\
\midrule
Federated Deep Learning$^{\dagger}$ & 3 & \emph{(P)} client drift in federated learning. \emph{(S)} batch normalization in federated learning; prototype-based federated learning \\
\midrule
Federated Machine Learning & 1 & \emph{(S)} vehicular federated learning \\
\midrule
Human--AI Interface & 2 & \emph{(P)} chatbot ethics. \emph{(S)} therapeutic conversational agents \\
\midrule
Large Language Models & 13 & \emph{(P)} LLM factuality; fairness in large language models; in-context learning mechanisms; knowledge-based visual question answering; reasoning abilities in large language models; zero-shot task generalization. \emph{(S)} biomedical domain-specific language models; code generation models; instruction-following benchmarks; knowledge graph integration in language models; long-context language models; retrieval-augmented question answering; retrieval-augmented reasoning \\
\midrule
Machine Unlearning & 6 & \emph{(P)} computationally efficient machine unlearning; privacy leakage in machine unlearning; training data protection from unauthorized model learning. \emph{(S)} approximate unlearning; data-free machine unlearning; privacy-preserving machine unlearning \\
\midrule
Masked Face Recognition & 2 & \emph{(S)} face occlusion reconstruction; transparent face masks$^{\ddagger}$ \\
\midrule
Masked Language Model & 2 & \emph{(P)} open-vocabulary object detection. \emph{(S)} contrastive representation learning \\
\midrule
Privacy-Preserving Machine Learning & 1 & \emph{(S)} graph neural networks \\
\midrule
Scientific Machine Learning & 4 & \emph{(P)} toxicity prediction. \emph{(S)} data-driven fluid modeling; differentiable simulation; machine-learning-assisted statistical inference \\
\midrule
Tiny Machine Learning & 1 & \emph{(S)} on-device gesture recognition \\
\midrule
Trustworthy AI & 2 & \emph{(P)} human--AI decision-making. \emph{(S)} sustainable AI \\
\midrule
Trustworthy Machine Learning$^{\dagger}$ & 6 & \emph{(P)} ethical issues in healthcare AI; shortcut learning. \emph{(S)} adversarial attacks on model explanations; contrastive learning; explainable clinical decision support systems; transparent machine learning \\
\midrule
Vertical Federated Learning & 2 & \emph{(S)} communication-efficient vertical federated learning; mutual information estimation in vertical federated learning \\
\midrule
\textbf{Total} & \textbf{59} & of 125 gate-passing candidates; the remaining \nweak{} are the validated weak signals of \ours \\
\bottomrule
\end{tabular}
\caption{Gate-passing candidates removed at expert validation, by mature target topic and signal space (\emph{P}: problem space, \emph{S}: solution space). The four automatic gates of \S\ref{sec:direction_focused} pass 125 candidates across the \ntopic{} target topics; two senior AI/ML researchers independently review all of them and retain \nweak{}, so the 59 listed here are the removals. Eight target topics do not appear: five produce no gate-passing candidate at all (\S\ref{app:absent}), and for the other three, \emph{Asynchronous Federated Learning}, \emph{Decentralized Federated Learning}, and \emph{Human-Centric AI}, every gate-passing candidate is retained. $^{\dagger}$The two topics whose every survivor is removed, leaving an empty signal set (\S\ref{app:absent}). $^{\ddagger}$The single removal on ground~(i), naming a physical object rather than a research direction; every other removal is on ground~(ii), a trajectory that lacks the low-then-exponentially-rising shape the definition requires (\S\ref{sec:backtrend-task}).}
\label{tab:removed-candidates}
\end{table}

\subsection{BERTScore Results}
\label{app:bertscore}
\paragraph{BERTScore-based Evaluation Details.}
For completeness, we additionally evaluate weak-signal discovery using two BERTScore-based settings, which complement the LLM-based metrics reported in the main text. Let the predicted weak-signal set be $s=\{s_1,\dots,s_m\}$ and the reference set be $g=\{g_1,\dots,g_n\}$. Unlike the LLM-based settings, which rely on explicit semantic judgments, the BERTScore-based settings use contextual embedding similarity to quantify semantic overlap.

In \emph{set-level BERTScore}, we first collapse each set into a single text sequence by concatenating all signals with a separator:
\begin{align*}
\mathrm{Concat}(s) &= s_1 \oplus s_2 \oplus \cdots \oplus s_m,\\
\mathrm{Concat}(g) &= g_1 \oplus g_2 \oplus \cdots \oplus g_n,
\end{align*}
where $\oplus$ denotes concatenation. We then compute BERTScore between the aggregated prediction and the aggregated reference:
\begin{multline*}
(P_{\text{set-bs}}, R_{\text{set-bs}}, F1_{\text{set-bs}}) \\
= \mathrm{BERTScore}\bigl(\mathrm{Concat}(s),\mathrm{Concat}(g)\bigr).
\end{multline*}
This setting measures coarse-grained similarity between the two sets as wholes, but does not explicitly model one-to-one or one-to-many correspondence between individual weak signals.

In \emph{signal-level BERTScore}, we instead compare weak signals individually. We first compute the full pairwise similarity matrix
\begin{align*}
S_{ij} &= \mathrm{BERTScore}_{F1}(s_i,g_j),\\
&\quad i=1,\dots,m,\; j=1,\dots,n,
\end{align*}
where each entry quantifies the semantic similarity between predicted signal $s_i$ and reference signal $g_j$. We then aggregate this matrix directionally. Precision is defined as the average best-match similarity from each prediction to the reference set:
\[
P_{\text{sig-bs}}
=
\frac{1}{m}\sum_{i=1}^{m}\max_{1 \leq j \leq n} S_{ij}.
\]
Intuitively, this asks: for each predicted signal, how well does its closest reference counterpart match? Recall is defined symmetrically as the average best-match similarity from each reference signal to the prediction set:
\[
R_{\text{sig-bs}}
=
\frac{1}{n}\sum_{j=1}^{n}\max_{1 \leq i \leq m} S_{ij}.
\]
This asks: for each reference weak signal, how well is it covered by the best prediction? We then combine the two using the harmonic mean:
\[
F1_{\text{sig-bs}}
=
\frac{2\,P_{\text{sig-bs}}R_{\text{sig-bs}}}
{P_{\text{sig-bs}}+R_{\text{sig-bs}}}.
\]
Compared with set-level BERTScore, this signal-level variant provides a finer-grained view of coverage by rewarding predictions that closely match individual reference weak signals rather than only the overall set semantics.

The main text reports only the LLM-judge and human-proxy settings; we give the two BERTScore-based settings here, together with the per-direction breakdown (\autoref{tab:appendix-bertscore-results}) and, below, an analysis of why BERTScore is uninformative for weak-signal matching.

  \begin{table}[htbp]
      \centering
      \small
      \setlength{\tabcolsep}{7pt}
      \renewcommand{\arraystretch}{1.15}
      \begin{tabular}{l ccc ccc}
      \toprule
      \multirow{2}{*}{\textbf{Model}} & \multicolumn{3}{c}{\textbf{Set-level BERTScore F1}} & \multicolumn{3}{c}{\textbf{Signal-level BERTScore F1}} \\
      \cmidrule(lr){2-4}\cmidrule(lr){5-7}
      & \textbf{Prob} & \textbf{Sol} & \textbf{All} & \textbf{Prob} & \textbf{Sol} & \textbf{All} \\
      \midrule
      GPT-5.4            & 81.7 & 82.9 & 82.3 & 86.9 & \textbf{88.3} & 87.6 \\
      Qwen3.5-397B       & 82.3 & 82.5 & 82.4 & 87.0 & 87.1 & 87.0 \\
      DeepSeek-R1-0528   & 82.3 & 82.9 & 82.6 & 87.1 & 87.7 & 87.4 \\
      Tongyi-DR-30B-A3B  & 83.1 & 82.9 & 83.0 & 87.5 & 87.4 & 87.4 \\
      DR-Tulu-8B         & 80.9 & 80.8 & 80.8 & 86.1 & 86.0 & 86.1 \\
      Qwen3-30B (RAG)    & \textbf{83.5} & \textbf{83.1} & \textbf{83.3} & \textbf{88.0} & 87.5 & \textbf{87.8} \\
      Qwen3-8B (RAG)     & 83.6 & 82.6 & 83.1 & 87.8 & 86.5 & 87.2 \\
      \bottomrule
      \end{tabular}
      \caption{Appendix BERTScore results on the current AI/ML benchmark, broken down by signal direction (\textbf{Prob}lem / \textbf{Sol}ution / \textbf{All}) for the set-level and signal-level settings. All values are percentages (\%). Higher is better; the best score in each column is in bold. All systems cluster within a few points, confirming that BERTScore is poorly discriminative for weak-signal matching (cf.\ the LLM-judge scores in \autoref{tab:main-results}). The spread across all seven systems is 2.5 points at the set level and 1.7 at the signal level, an order of magnitude narrower than the LLM-judge spread on the same predictions.}
      \label{tab:appendix-bertscore-results}
  \end{table}

\paragraph{Why BERTScore is high and non-discriminative.}
Signal-level BERTScore aggregates token-level cosine similarities between contextual roberta-large embeddings via greedy best-match matching. Every weak signal in \ours is a short noun phrase from a single domain, assembled from a small shared vocabulary: \emph{learning}, \emph{model}, \emph{language}, \emph{federated}, \emph{augmented}, \emph{neural}, \emph{training}, \emph{transformer}. Two such phrases therefore share many high-similarity tokens \emph{regardless} of whether they denote the same research direction, and roberta-large places domain terms close together in embedding space; without baseline rescaling, even unrelated same-domain phrases score above $0.85$. BERTScore thus measures \emph{``are these both AI/ML phrases''} rather than \emph{``are these the same specific precursor''}, which is exactly the distinction the benchmark hinges on. This is why all seven systems cluster at $81$--$88$ (\autoref{tab:appendix-bertscore-results}) irrespective of correctness.

\paragraph{BERTScore's ``matches'' are frequently the wrong research direction.}
\autoref{tab:bertscore-false} makes the failure concrete: for each reference signal we show the prediction that BERTScore scores \emph{highest}, and it is repeatedly a different research direction that the LLM judge (and inspection) reject. These false matches score $86.7$--$91.2$, on par with or above genuine matches. Most strikingly, for the reference \emph{long-context understanding in large language models} BERTScore ranks the unrelated \emph{interpretability and mechanistic understanding deficits} ($91.2$) \emph{above} the correct \emph{long-context dependence and context window limitations} ($90.1$), and for \emph{retrieval-augmented language models} it ranks \emph{program-aided and code-interpreted reasoning} ($89.8$) above the correct \emph{retrieval-augmented generation} ($89.6$). In both cases a shared abstract-noun template --- ``\dots understanding \dots'' or ``\dots-aided/-augmented \dots reasoning'' --- dominates the token overlap, while the one word that carries the meaning barely moves the score. A metric that ranks a wrong precursor above the right one cannot separate systems on this task, which is why we treat the LLM judge, whose near-floor verdict a judge from a third model family corroborates in \autoref{tab:main-results}, as the primary evaluator and relegate BERTScore to this appendix.

\begin{table}[htbp]
  \centering
  \small
  \renewcommand{\arraystretch}{1.15}
  \setlength{\tabcolsep}{5pt}
  \begin{tabular}{>{\raggedright\arraybackslash}p{0.30\linewidth} >{\raggedright\arraybackslash}p{0.32\linewidth} c >{\raggedright\arraybackslash}p{0.20\linewidth}}
  \toprule
  \textbf{Reference weak signal} & \textbf{Prediction BERTScore ranks highest} & \textbf{F1} & \textbf{Why it is not a match} \\
  \midrule
  parameter-efficient tuning & Parameter-efficient finetuning & 92.0 & \emph{(the correct match, shown for reference)} \\
  long-context understanding in large language models & Interpretability and mechanistic understanding deficits & \textbf{91.2} & interpretability $\neq$ long context, and it outranks the correct \emph{Long-context dependence and context window limitations} (90.1) \\
  retrieval-augmented language models & Program-aided and code-interpreted reasoning & \textbf{89.8} & code execution $\neq$ retrieval, and it outranks the correct \emph{Retrieval-augmented generation} (89.6) \\
  post-training quantization for large language models & Tool use via language-model orchestration & 89.3 & model compression $\neq$ agentic tool use \\
  privacy leakage in NLP models & Benchmark contamination and data leakage & 88.9 & test-set contamination $\neq$ leaking training data; only the word ``leakage'' overlaps \\
  machine-generated text detection & In-context generalization without reliable evaluation & 87.6 & detecting generated text $\neq$ few-shot generalization \\
  LLM-based automated program repair & Scaling-law-guided model development & 86.7 & code repair $\neq$ scaling laws \\
  \bottomrule
  \end{tabular}
  \caption{BERTScore false matches on the \emph{large language models} topic (GPT-5.4 predictions, both signal directions). For each reference weak signal we list the prediction that signal-level BERTScore scores highest and its F1 (\%). The first row is the genuine match, included as a baseline. Every other row is a different research direction that the LLM judge rejects, yet all of them score between $86.7$ and $91.2$, indistinguishable from that baseline. In two cases the top-scoring prediction is not merely wrong but outranks the correct one: the judge accepts \emph{Retrieval-augmented generation} and \emph{Long-context dependence and context window limitations}, and BERTScore places a rejected prediction above each. BERTScore rewards shared domain vocabulary and phrase templates rather than same-topic identity.}
  \label{tab:bertscore-false}
\end{table}

\subsection{Additional Error Cases}
  \label{app:error-cases}

  This appendix expands the discussion in Section~\ref{sec:error-analysis} by providing concrete benchmark--prediction
  comparisons. The goal is to show that many low-scoring cases are not random failures, but systematic mismatches in framing,
  abstraction level, or semantic coverage. In particular, several examples are highly plausible on their own terms, which helps
  explain why lexical similarity can remain high even when semantic judge-based matching is low.

  The examples in \autoref{tab:error-cases-appendix} follow the same four error types discussed in the main text. First, some
  cases exhibit \emph{topic drift}, where the model stays within a reasonable neighboring area but shifts away from the
  benchmark's intended problem definition. Second, some cases show \emph{granularity mismatch}, where the model predicts narrower
  mechanisms or engineering details while the benchmark is formulated at the level of research paradigms or system bottlenecks.
  Third, some examples are \emph{lexical near-misses}: the predictions are clearly in the same field and share substantial
  vocabulary with the benchmark, but the actual research framing is different. Finally, some cases demonstrate \emph{coverage
  failure}, where the model captures one dimension of the benchmark but omits other equally central dimensions.

  \begin{table}[htbp]
    \centering
    \scriptsize
    \renewcommand{\arraystretch}{1.15}
    \setlength{\tabcolsep}{3pt}
    \begin{tabular}{
      >{\raggedright\arraybackslash}p{0.14\linewidth}
      >{\raggedright\arraybackslash}p{0.08\linewidth}
      >{\raggedright\arraybackslash}p{0.23\linewidth}
      >{\raggedright\arraybackslash}p{0.23\linewidth}
      >{\raggedright\arraybackslash}p{0.23\linewidth}
    }
    \toprule
    \textbf{Topic} & \textbf{Direction} & \textbf{Reference signal(s)} & \textbf{GPT-5.4 prediction(s)} \textbf{(rank)} & \textbf{Error interpretation} \\
    \midrule

    Trustworthy AI
    & Problem
    & AI trust measurement; appropriate reliance on AI; AI trust calibration; ethical AI in chatbot systems
    & Robustness under distribution shift (1); uncertainty calibration (2); out-of-distribution detection (3); spurious correlation dependence (4)
    & Topic drift from human-facing trust calibration and measurement to generic model-robustness and safety problems. \emph{Uncertainty calibration} is the nearest miss, but calibrating a model's confidence is not calibrating a user's trust; none of the 13 ranked predictions matches. \\
    \midrule

    Attention Mechanisms in CNN
    & Solution
    & Linear-complexity self-attention; sparse attention in vision transformers; large-kernel convolutions; locality-enhanced transformer architectures; hierarchical vision transformers
    & Lightweight channel--spatial gating modules (1); multi-scale feature reweighting (2); dynamic convolution via input-conditioned kernel weighting (3); efficient attention approximations for vision backbones (9)
    & Granularity mismatch: the top of the ranking is fine-grained attention-block variants rather than the benchmark's architecture-family solutions. The single match, \emph{efficient attention approximations}, surfaces only at rank 9. \\
    \midrule

    Masked Language Model
    & Solution
    & Vision-language models; prompt tuning for pretrained language models; masked video modeling
    & Span corruption denoising objectives (1); replaced-token detection objectives (2); dynamic masking during pretraining (3); multimodal masking objectives (12)
    & Lexical near-miss: all 14 predictions stay inside masked-LM pretraining objectives. The two that do point outward, \emph{multimodal masking objectives} and \emph{retrieval-augmented encoder representations}, sit at ranks 12 and 14 and still miss the downstream paradigms the benchmark tracks. \\
    \midrule

    Vertical Federated Learning
    & Problem
    & Straggler problem; privacy leakage; feature selection; unaligned data (all in vertical FL)
    & Privacy-preserving entity resolution across siloed parties (1); partial-overlap sample populations (3); leakage risk from intermediate predictions and gradients (5); label leakage from party roles and training interactions (10)
    & Coverage failure on the highest-scoring pair in the benchmark (F1 $41.8$): unaligned data and privacy leakage are matched at ranks 3 and 5, yet the top-ranked prediction reframes alignment as entity resolution and feature selection is absent from all 13. \\
    \midrule

    Machine Unlearning
    & Problem
    & Federated unlearning; graph unlearning
    & Certified removal of data influence (2); retraining cost after data deletion (3); cross-silo deletion in federated or distributed training (9)
    & Coverage failure: the federated setting is recovered only at rank 9, graph unlearning is absent from all 12 predictions, and the top ranks are generic erasure-compliance concerns. \\
    \midrule

    Large Language Models
    & Problem
    & Privacy leakage in NLP models; long-context understanding; LLM alignment with human feedback; machine-generated text detection; pre-trained LM compression
    & Hallucination and factual unreliability (2); benchmark contamination and data leakage (3); long-context dependence and context window limitations (10); memorization, privacy leakage, and unintended training-data reproduction (11)
    & Coverage failure: two of the five references are recovered, but only at ranks 10--11; text detection and model compression are absent from all 16 predictions, and the first nine ranks are evaluation and reliability concerns outside the reference set. \\
    \bottomrule
    \end{tabular}
    \caption{Representative benchmark--prediction comparisons for GPT-5.4 predictions on the AI/ML target topics. Predictions are quoted from the ranked lists scored in \autoref{tab:main-results}, with each one's rank in the system's own confidence ordering in parentheses. The examples illustrate that many low-scoring outputs are semantically plausible near-misses that diverge in framing, abstraction level, or benchmark coverage. The ranks also show where the misalignment sits: when a reference signal is recovered at all, it is usually recovered late.}
    \label{tab:error-cases-appendix}
  \end{table}
  
\FloatBarrier

\subsection{Prompts}
\label{prompt}
In this section, we present the comprehensive prompts that we use to mine candidate topics (\autoref{fig:construct-prompt-a}, continued in \autoref{fig:construct-prompt-b}), elicit weak-signal predictions from the evaluated systems (\autoref{fig:pred-problem-prompt}, \autoref{fig:pred-solution-prompt}), and conduct LLM-as-a-judge evaluations (\autoref{fig:set-judge-prompt}, \autoref{fig:signal-judge-prompt}). We use Artificial Intelligence and Machine Learning as our field, and the prompts can be easily adapted for other fields.

\begin{center}
\begin{figure*}[p]
  \centering
  \begin{tcolorbox}[
    colback=black!7.5!white,
    colframe=black!80!white,
    title={Construction Prompt: Candidate Topic Extraction (System Prompt, and User Prompt Part 1)},
    fontupper=\scriptsize,
    fonttitle=\footnotesize,
    boxrule=0.4pt,
    arc=0.8mm,
    left=0.8mm,
    right=0.8mm,
    top=0.6mm,
    bottom=0.6mm
  ]
  \setlength{\parskip}{0pt}\raggedright
  {\textbf{System prompt.}} \\
  {You are an expert research topic extractor.} \\
  {Your task is to extract literature-level research topics from paper abstracts.} \\
  {Extract only topics that are explicitly discussed in the paper.} \\
  {A good topic should be broad enough that multiple independent papers could study it,} \\
  {but specific enough to be more informative than a general field label.} \\
  {Prefer clean, compact topic labels that name the core research problem or method family.} \\
  {Return only valid json.} \\[7pt]
  {\textbf{User prompt.}} \\
  {\textbf{Target established topic:}} \\
  {\{target\_topic\}} \\[5pt]
  {This paper was retrieved as related to the target established topic above.} \\
  {Use the target topic only as context for relevance filtering.} \\
  {Do not output the target topic itself unless the abstract discusses a more specific reusable subtopic.} \\[5pt]
  {\textbf{Paper metadata:}} \\
  {- Title: \{title\}} \\
  {- Paper ID: \{paper\_id\}} \\
  {- Year: \{year\}} \\
  {- Venue: \{venue\}} \\
  {- Source query: \{source\_query\}} \\[5pt]
  {\textbf{Abstract:}} \\
  {\{abstract\}} \\[5pt]
  {\textbf{Topic categories:}} \\[5pt]
  {\textbf{1. Problem-space topics:}} \\
  {Research problems, gaps, limitations, risks, bottlenecks, evaluation failures, or scientific questions discussed by the paper.} \\[5pt]
  {\textbf{2. Solution-space topics:}} \\
  {Research methods, method families, system directions, evaluation approaches, defenses, or solution directions discussed by the paper.} \\
  {Use solution-space only for standalone reusable methods, method families, systems, defenses, algorithms, datasets, benchmarks, or evaluation protocols, not for the problem that motivates them.} \\[5pt]
  {\textbf{Task:}} \\
  {Extract clean, reusable research topics from this paper abstract that are conceptually related to the target established topic: ``\{target\_topic\}''.} \\
  {These candidates may be problem-space topics or solution-space topics.} \\
  {Most abstracts describe both a problem/gap and a method/solution. Separate these roles.} \\
  {Do not combine a method, remedy, evaluation detail, dataset, application setting, or implementation detail with the problem it addresses.} \\[5pt]
  {\textbf{Specificity guidance:}} \\
  {- Too broad: a whole field (e.g., ``machine learning'', ``computer vision''), broad model family (e.g., ``deep learning''), or generic category label (e.g., ``optimization'').} \\
  {- Too specific: a paper-specific method name (e.g., ``CoPINet''), system name (e.g., ``GPT Semantic Cache''), exact task setting (e.g., ``Raven's Progressive Matrices''), implementation detail, single experimental finding (e.g., ``68.8\% API-call reduction''), or enumerating technical details (e.g., ``using interventional data'', ``with linguistically regularized CNN'', ``additive feature attribution'').} \\
  {- Correct level: a reusable research direction or problem space that multiple independent papers could study using different methods, or systems. The topic should capture the core research direction without enumerating specific techniques or data types.} \\
  {- Prefer compact labels such as ``causal model evaluation'' over long contribution phrases such as ``evaluation of causal models using interventional empirical data''.} \\
  {- Focus on the research problem or method family, not on the specific implementation details or data modalities.} \\
  {- If the abstract discusses a narrow technique or case study, abstract it to the broader research problem or method family it addresses.} \\
  {- Do not phrase topics as actions (e.g., ``evaluation of'', ``analysis of'') or as this specific paper's contribution.} \\
  {- A candidate topic must be a standalone research topic phrase, not a relation between a problem and a solution.} \\
  {- Avoid ``X for Y'' topic names when X is a method and Y is a problem, goal, task, or desired property. Split them into separate candidate topics when both are explicitly supported.} \\
  {- If the paper proposes method X to solve, evaluate, improve, or measure problem Y, output Y as a problem-space topic and X as a solution-space topic only when each is independently reusable. Do not output the combined phrase.}
  \end{tcolorbox}
  \caption{{Construction prompt used to extract problem- and solution-space candidate topics from each retrieved paper abstract, reproduced verbatim from the released pipeline code; braces mark the template slots filled in at run time: the full system prompt, and the user prompt up to the specificity guidance. The remainder is shown in \autoref{fig:construct-prompt-b}.}}
  \label{fig:construct-prompt-a}
  \end{figure*}

\end{center}

\begin{center}
\begin{figure*}[p]
  \centering
  \begin{tcolorbox}[
    colback=black!7.5!white,
    colframe=black!80!white,
    title={Construction Prompt: Candidate Topic Extraction (User Prompt Part 2)},
    fontupper=\scriptsize,
    fonttitle=\footnotesize,
    boxrule=0.4pt,
    arc=0.8mm,
    left=0.8mm,
    right=0.8mm,
    top=0.6mm,
    bottom=0.6mm
  ]
  \setlength{\parskip}{0pt}\raggedright
  {\textbf{Bad topic examples (wrong abstraction level):}} \\
  {- ``machine learning'' because it is too broad} \\
  {- ``Empirical evaluation of causal models using interventional data'' because it enumerates technical details (``using interventional data'')} \\
  {- ``evaluation of causal modeling algorithms using interventional empirical data'' because it mixes the core problem with proposed evaluation details; prefer ``causal model evaluation'' as the problem-space topic.} \\
  {- ``CoPINet for Raven's Progressive Matrices'' because it is paper-specific} \\
  {- ``Aspect-based sentiment analysis with linguistically regularized CNN'' because it enumerates method details} \\
  {- ``Explainability challenges in additive feature attribution'' because it is too method-specific} \\
  {- ``deep reinforcement learning for federated edge learning resource management'' because it combines a solution method with a problem/context; split into cleaner topics only if each side is independently reusable and explicitly supported.} \\
  {- ``Using interaction-aware explanations to solve misleading model interpretability'' because it mixes a solution and a problem in one extracted topic; extract the problem topic and solution topic separately if both are explicitly supported.} \\
  {- ``uncertainty communication for AI trust calibration'' because it mixes a solution method with the problem or goal it addresses; extract ``AI trust calibration'' as a problem-space topic or ``uncertainty communication in AI systems'' as a solution-space topic if each is explicitly supported.} \\[5pt]
  {\textbf{Requirements:}} \\
  {- Output a json object with a ``topics'' array, which may be empty.} \\
  {- Extract up to 2 topics total.} \\
  {- Each topic must be explicitly supported by the abstract.} \\
  {- Each topic must be reusable across multiple papers.} \\
  {- Each topic must be one abstraction level broader than the paper's specific method, benchmark, dataset, or case study.} \\
  {- Each topic must include exactly one ``topic\_type'': ``problem-space'' or ``solution-space''. A mix is not allowed.} \\
  {- Each topic must be conceptually related to the target established topic: ``\{target\_topic\}''.} \\
  {- If a topic cannot be clearly classified as problem-space or solution-space, do not emit it.} \\
  {- If a topic cannot be clearly related to the target established topic, do not emit it.} \\
  {- Do not phrase topics as actions, paper contributions, or problem-solution relationships.} \\
  {- Do not combine a method and the problem it addresses into one topic phrase.} \\
  {- Do not include proposed measurement choices, dataset choices, application settings, or implementation details in the topic label unless they are themselves the reusable research topic.} \\
  {- Do not invent evidence beyond the abstract.} \\[5pt]
  {\textbf{Return only json with this schema:}} \\
  \texttt{\{} \\
  \texttt{\hspace*{1.1em}"topics": [} \\
  \texttt{\hspace*{2.2em}\{} \\
  \texttt{\hspace*{3.3em}"topic": "<standalone literature-level candidate topic>",} \\
  \texttt{\hspace*{3.3em}"topic\_type": "problem-space|solution-space",} \\
  \texttt{\hspace*{3.3em}"target\_topic": "\{target\_topic\}",} \\
  \texttt{\hspace*{3.3em}"evidence": "<short phrase grounded in the abstract>",} \\
  \texttt{\hspace*{3.3em}"confidence": "high|medium|low"} \\
  \texttt{\hspace*{2.2em}\}} \\
  \texttt{\hspace*{1.1em}]} \\
  \texttt{\}}
  \end{tcolorbox}
  \caption{{Continuation of the candidate-topic extraction user prompt (\autoref{fig:construct-prompt-a}): bad-topic examples, output requirements, and the response schema. Extracted candidates are then deduplicated, embedding-clustered, and scored by early frequency and 2024 reference adoption before expert validation (\S\ref{sec:direction_focused}).}}
  \label{fig:construct-prompt-b}
  \end{figure*}

\end{center}

\begin{center}
\begin{figure*}[p]
  \centering
  \begin{tcolorbox}[
    colback=black!7.5!white,
    colframe=black!80!white,
    title={Prediction Prompt for Problem-Space Weak Signals},
    fontupper=\scriptsize,
    fonttitle=\footnotesize,
    boxrule=0.4pt,
    arc=0.8mm,
    left=0.8mm,
    right=0.8mm,
    top=0.6mm,
    bottom=0.6mm
  ]
  \setlength{\parskip}{0pt}\raggedright
  {You are an expert analyst of frontier \{domain\} research. Your task is to identify early weak signals that later contributed to a specified mature target topic.} \\[5pt]
  {We distinguish two categories of weak signals. One category is the ``solution-space weak signal'': an early research method, technique, or design principle that was not yet widely adopted but later became an important solution to an already-recognized problem. } \\[5pt]
  {However, the category you are asked to identify here is the ``problem-space weak signal'': an underrecognized research problem or problem formulation that was not yet widely recognized by the research community at the time it emerged, but that later became central to the mature target topic [\{mainframe\_topic\}].} \\
  {To be more specific, problem-space weak signals include research problems, gaps, limitations, risks, bottlenecks, evaluation failures, or scientific questions that emerged in the literature during the prediction window, rather than claims tied to a single paper.} \\[5pt]
  {\textbf{Mature target topic:}} \\
  {[\{mainframe\_topic\}]} \\[5pt]
  {\textbf{Prediction window:}} \\
  {[\{year\_range\}]} \\[5pt]
  {\textbf{Retrospective setup:}} \\
  {The mature target topic should be treated as a topic that is already established or prominent by 2024. Your task is not to predict after 2024. Your task is to look backward and identify what this 2024 mature topic looked like in the 2019-2023 literature while it was still emerging.} \\
  {Use the mature target topic only as 2024 relevance context. The weak signals themselves must be research problems, gaps, limitations, risks, bottlenecks, or scientific questions that appeared in [\{year\_range\}]. Do not use 2024-or-later evidence, and do not simply restate the mature target topic unless you name a more specific predecessor formulation.} \\[5pt]
  {\textbf{Question:}} \\
  {What are the early problem-space weak signals in \{domain\} that emerged between [\{year\_range\}] and later contributed to the mature target topic [\{mainframe\_topic\}]?} \\[5pt]
  {\textbf{Specificity guidance:}} \\
  {- Use the mature target topic only as relevance context.} \\
  {- Do not output the target topic itself unless you name a more specific reusable subtopic or predecessor direction.} \\
  {- Too broad: a whole field (e.g., ``machine learning'', ``computer vision''), broad model family (e.g., ``deep learning''), or generic category label (e.g., ``optimization'').} \\
  {- Too specific: a paper-specific method name, system name, exact dataset, benchmark instance, implementation detail, single experimental finding, or single case study.} \\
  {- Correct level: a reusable research direction or problem space topic that multiple independent papers could study using different methods or systems.} \\
  {- Focus on the research problem, not on specific implementation details or data modalities.} \\
  {- Do not phrase signals as actions, paper contributions, or problem-solution relationships.} \\
  {- Avoid ``X for Y'' signal names when X is a method and Y is a problem, goal, task, or desired property. For problem-space output, keep only the problem side (i.e., Y) when it is independently reusable and relevant.} \\[5pt]
  {\textbf{Requirements:}} \\
  {- Return ONLY valid JSON, no markdown fences, no explanation.} \\
  {- Output a JSON object with a ``weak\_signals'' array. Order the ``weak\_signals'' array from most to least confident: the array order is your ranking, strongest first.} \\
  {- Output at least 10 weak signals unless you genuinely cannot name that many.} \\
  {- Each weak signal must have exactly these fields: ``signal'', ``what\_it\_was'', ``why\_weak\_signal''.} \\
  {- Each weak signal must be explicitly tied to the prediction window [\{year\_range\}].} \\
  {- ``what\_it\_was'' must include the year or year range within [\{year\_range\}].} \\
  {- ``why\_weak\_signal'' must explain why this was a problem-space weak signal for [\{mainframe\_topic\}].} \\
  {- Each signal must be conceptually related to [\{mainframe\_topic\}].} \\
  {- Each signal must be reusable across multiple papers.} \\
  {- Do not include solution methods.} \\
  {- Do not invent evidence or overclaim certainty.} \\[5pt]
  {\textbf{Return only JSON with this schema:}} \\
  \texttt{\{} \\
  \texttt{\hspace*{1.1em}"weak\_signals": [} \\
  \texttt{\hspace*{2.2em}\{} \\
  \texttt{\hspace*{3.3em}"signal": "<short reusable problem-space weak signal name>",} \\
  \texttt{\hspace*{3.3em}"what\_it\_was": "<1-2 sentences describing what it was, including the year>",} \\
  \texttt{\hspace*{3.3em}"why\_weak\_signal": "<1-2 sentences explaining why it was a problem-space weak signal for [\{mainframe\_topic\}]>"} \\
  \texttt{\hspace*{2.2em}\}} \\
  \texttt{\hspace*{1.1em}]} \\
  \texttt{\}}
  \end{tcolorbox}
  \caption{{Prediction prompt template for problem-space weak signals, reproduced verbatim from the released pipeline code; braces mark the template slots filled in at run time. \texttt{domain} is Artificial Intelligence and Machine Learning, \texttt{mainframe\_topic} the mature target topic $M$, and \texttt{year\_range} the 2019--2023 prediction window. Every evaluated system receives this as a single user message with no system prompt; the RAG systems additionally append the retrieved evidence block and the instruction ``Use this evidence when generating the weak signals.'' (\S\ref{sec:evaluated-systems}).}}
  \label{fig:pred-problem-prompt}
  \end{figure*}

\end{center}

\begin{center}
\begin{figure*}[p]
  \centering
  \begin{tcolorbox}[
    colback=black!7.5!white,
    colframe=black!80!white,
    title={Prediction Prompt for Solution-Space Weak Signals},
    fontupper=\scriptsize,
    fonttitle=\footnotesize,
    boxrule=0.4pt,
    arc=0.8mm,
    left=0.8mm,
    right=0.8mm,
    top=0.6mm,
    bottom=0.6mm
  ]
  \setlength{\parskip}{0pt}\raggedright
  {You are an expert analyst of frontier \{domain\} research. Your task is to identify early weak signals that later contributed to a specified mature target topic.} \\[5pt]
  {We distinguish two categories of weak signals. One category is the ``problem-space weak signal'': an underrecognized research problem or problem formulation that was not yet widely recognized by the research community at the time it emerged, but that later became central to the mature target topic [\{mainframe\_topic\}].} \\[5pt]
  {However, the category you are asked to identify here is the ``solution-space weak signal'': an early research method, technique, or design principle that was not yet widely adopted but later became an important solution to an already-recognized problem.} \\
  {To be more specific, solution-space weak signals include research methods, method families, system directions, evaluation approaches, defenses, or solution directions that emerged in the literature during the prediction window, rather than claims tied to a single paper.} \\[5pt]
  {\textbf{Mature target topic:}} \\
  {[\{mainframe\_topic\}]} \\[5pt]
  {\textbf{Prediction window:}} \\
  {[\{year\_range\}]} \\[5pt]
  {\textbf{Retrospective setup:}} \\
  {The mature target topic should be treated as a topic that is already established or prominent by 2024. Your task is not to predict after 2024. Your task is to look backward and identify what this 2024 mature topic looked like in the 2019-2023 literature while it was still emerging.} \\
  {Use the mature target topic only as 2024 relevance context. The weak signals themselves must be research methods, method families, system directions, evaluation approaches, defenses, or solution directions that appeared in [\{year\_range\}]. Do not use 2024-or-later evidence, and do not simply restate the mature target topic unless you name a more specific predecessor formulation.} \\[5pt]
  {\textbf{Question:}} \\
  {What are the early solution-space weak signals in \{domain\} that emerged between [\{year\_range\}] and later contributed to the mature target topic [\{mainframe\_topic\}]?} \\[5pt]
  {\textbf{Specificity guidance:}} \\
  {- Use the mature target topic only as relevance context.} \\
  {- Do not output the target topic itself unless you name a more specific reusable subtopic or predecessor direction.} \\
  {- Too broad: a whole field (e.g., ``machine learning'', ``computer vision''), broad model family (e.g., ``deep learning''), or generic category label (e.g., ``optimization'').} \\
  {- Too specific: a paper-specific method name, system name, exact dataset, benchmark instance, implementation detail, single experimental finding, or single case study.} \\
  {- Correct level: a reusable research method, method family, system direction, evaluation approach, defense, or solution direction that multiple independent papers could study.} \\
  {- Focus on the research method or solution direction, not on specific implementation details, problem formulations, or data modalities.} \\
  {- Do not phrase signals as actions, paper contributions, or problem-solution relationships.} \\
  {- Avoid ``X for Y'' signal names when X is a method and Y is a problem, goal, task, or desired property. For solution-space output, keep only the solution side (i.e., X) when it is independently reusable and relevant.} \\[5pt]
  {\textbf{Requirements:}} \\
  {- Return ONLY valid JSON, no markdown fences, no explanation.} \\
  {- Output a JSON object with a ``weak\_signals'' array. Order the ``weak\_signals'' array from most to least confident: the array order is your ranking, strongest first.} \\
  {- Output at least 10 weak signals unless you genuinely cannot name that many.} \\
  {- Each weak signal must have exactly these fields: ``signal'', ``what\_it\_was'', ``why\_weak\_signal''.} \\
  {- Each weak signal must be explicitly tied to the prediction window [\{year\_range\}].} \\
  {- ``what\_it\_was'' must include the year or year range within [\{year\_range\}].} \\
  {- ``why\_weak\_signal'' must explain why this was a solution-space weak signal for [\{mainframe\_topic\}].} \\
  {- Each signal must be conceptually related to [\{mainframe\_topic\}].} \\
  {- Each signal must be reusable across multiple papers.} \\
  {- Do not include problem statements.} \\
  {- Do not invent evidence or overclaim certainty.} \\[5pt]
  {\textbf{Return only JSON with this schema:}} \\
  \texttt{\{} \\
  \texttt{\hspace*{1.1em}"weak\_signals": [} \\
  \texttt{\hspace*{2.2em}\{} \\
  \texttt{\hspace*{3.3em}"signal": "<short reusable solution-space weak signal name>",} \\
  \texttt{\hspace*{3.3em}"what\_it\_was": "<1-2 sentences describing what it was, including the year>",} \\
  \texttt{\hspace*{3.3em}"why\_weak\_signal": "<1-2 sentences explaining why it was a solution-space weak signal for [\{mainframe\_topic\}]>"} \\
  \texttt{\hspace*{2.2em}\}} \\
  \texttt{\hspace*{1.1em}]} \\
  \texttt{\}}
  \end{tcolorbox}
  \caption{{Prediction prompt template for solution-space weak signals, reproduced verbatim from the released pipeline code; braces mark the template slots filled in at run time. \texttt{domain} is Artificial Intelligence and Machine Learning, \texttt{mainframe\_topic} the mature target topic $M$, and \texttt{year\_range} the 2019--2023 prediction window. Every evaluated system receives this as a single user message with no system prompt; the RAG systems additionally append the retrieved evidence block and the instruction ``Use this evidence when generating the weak signals.'' (\S\ref{sec:evaluated-systems}).}}
  \label{fig:pred-solution-prompt}
  \end{figure*}

\end{center}

\begin{center}
\begin{figure*}[p]
  \centering
  \begin{tcolorbox}[
    colback=black!7.5!white,
    colframe=black!80!white,
    title={Set-Level LLM-as-a-Judge Prompt},
    fontupper=\scriptsize,
    fonttitle=\footnotesize,
    boxrule=0.4pt,
    arc=0.8mm,
    left=0.8mm,
    right=0.8mm,
    top=0.6mm,
    bottom=0.6mm
  ]
  \setlength{\parskip}{0pt}\raggedright
  {\textbf{System prompt.}} \\
  {You are a research topic matcher. Two research topics match only if they denote the same specific research topic: the same core method or the same problem, differing at most in wording, phrasing, or abbreviation. Topics that are merely related, adjacent, complementary, or from the same broad area do not match.} \\[7pt]
  {\textbf{User prompt.}} \\
  {Compare the PREDICTED set against the GROUND TRUTH set of research topics.} \\[5pt]
  {Precision = fraction of predicted topics that match some ground-truth topic.} \\
  {Recall = fraction of ground-truth topics that match some predicted topic.} \\[5pt]
  {\textbf{Ground truth:}} \\
  {\{gt\_block\}} \\[5pt]
  {\textbf{Predicted:}} \\
  {\{pred\_block\}} \\[5pt]
  {Return ONLY valid JSON:} \\
  {\{``precision'': <float 0-1>, ``recall'': <float 0-1>, ``matched\_pairs'': [\{``gt\_index'': <int>, ``pred\_index'': <int>\}]\}} \\
  {Give one matched\_pairs entry per matching (ground-truth, predicted) pair; empty list if none.}
  \end{tcolorbox}
  \caption{{Set-level LLM-as-a-judge prompt, reproduced verbatim from the released pipeline code; braces mark the template slots filled in at run time. Both judges share the same system prompt and differ only in the user message. \texttt{gt\_block/pred\_block} are the reference and prediction lists, rendered as 0-based numbered items. A match requires the two topics to denote the same specific research topic, not merely a related one (\S\ref{sec:eval-protocol}).}}
  \label{fig:set-judge-prompt}
  \end{figure*}

\end{center}

\begin{center}
\begin{figure*}[p]
  \centering
  \begin{tcolorbox}[
    colback=black!7.5!white,
    colframe=black!80!white,
    title={Signal-Level LLM-as-a-Judge Prompt},
    fontupper=\scriptsize,
    fonttitle=\footnotesize,
    boxrule=0.4pt,
    arc=0.8mm,
    left=0.8mm,
    right=0.8mm,
    top=0.6mm,
    bottom=0.6mm
  ]
  \setlength{\parskip}{0pt}\raggedright
  {\textbf{System prompt.}} \\
  {You are a research topic matcher. Two research topics match only if they denote the same specific research topic: the same core method or the same problem, differing at most in wording, phrasing, or abbreviation. Topics that are merely related, adjacent, complementary, or from the same broad area do not match.} \\[7pt]
  {\textbf{User prompt.}} \\
  {For each candidate research topic, decide whether it matches any reference research topic.} \\[5pt]
  {\textbf{Reference:}} \\
  {\{ref\_block\}} \\[5pt]
  {\textbf{Candidate:}} \\
  {\{cand\_block\}} \\[5pt]
  {Return ONLY valid JSON: \{``matches'': [<int>, ...]\} with exactly \{n\_cand\} elements, 1 if the candidate matches any reference else 0.}
  \end{tcolorbox}
  \caption{{Signal-level LLM-as-a-judge prompt, reproduced verbatim from the released pipeline code; braces mark the template slots filled in at run time. Both judges share the same system prompt and differ only in the user message. \texttt{ref\_block/cand\_block} are the reference and prediction lists, rendered as 0-based numbered items. A match requires the two topics to denote the same specific research topic, not merely a related one (\S\ref{sec:eval-protocol}).}}
  \label{fig:signal-judge-prompt}
  \end{figure*}

\end{center}

\end{document}